\documentclass[final]{cvpr}

\usepackage{times}
\usepackage{epsfig}
\usepackage{graphicx}
\usepackage{amsmath}
\usepackage{amssymb}
\usepackage{algorithm}
\usepackage{algorithmic}
\usepackage{mathbbol}
\usepackage{caption}
\newcommand{\1}{\mbox{1}\hspace{-0.25em}\mbox{l}}

\usepackage{color}
\newcommand{\chkA}[1]{#1}
\newcommand{\chkB}[1]{#1}
\newcommand{\chkC}[1]{#1}
\newcommand{\chkD}[1]{#1}
\newcommand{\chkE}[1]{#1}
\newcommand{\chkF}[1]{#1}
\newcommand{\chkG}[1]{#1}
\newcommand{\chkURL}[1]{\textcolor{magenta}{#1}}

\newcommand{\best}[1]{#1}
\newcommand{\sbest}[1]{#1}

\usepackage[pagebackref=true,breaklinks=true,colorlinks,bookmarks=false]{hyperref}
\usepackage{breakurl}

\def\cvprPaperID{} %
\def\confYear{}

\begin{document}

\title{Few-shot Semantic Image Synthesis Using StyleGAN Prior}

\author{Yuki Endo\\
University of Tsukuba\\
{\tt\small endo@cs.tsukuba.ac.jp}
\and
Yoshihiro Kanamori\\
Univeristy of Tsukuba\\
{\tt\small kanamori@cs.tsukuba.ac.jp}
}

\twocolumn[{%
\renewcommand\twocolumn[1][]{#1}%
\maketitle
\begin{center}
    \centering
    \includegraphics[width=1.\linewidth, clip]{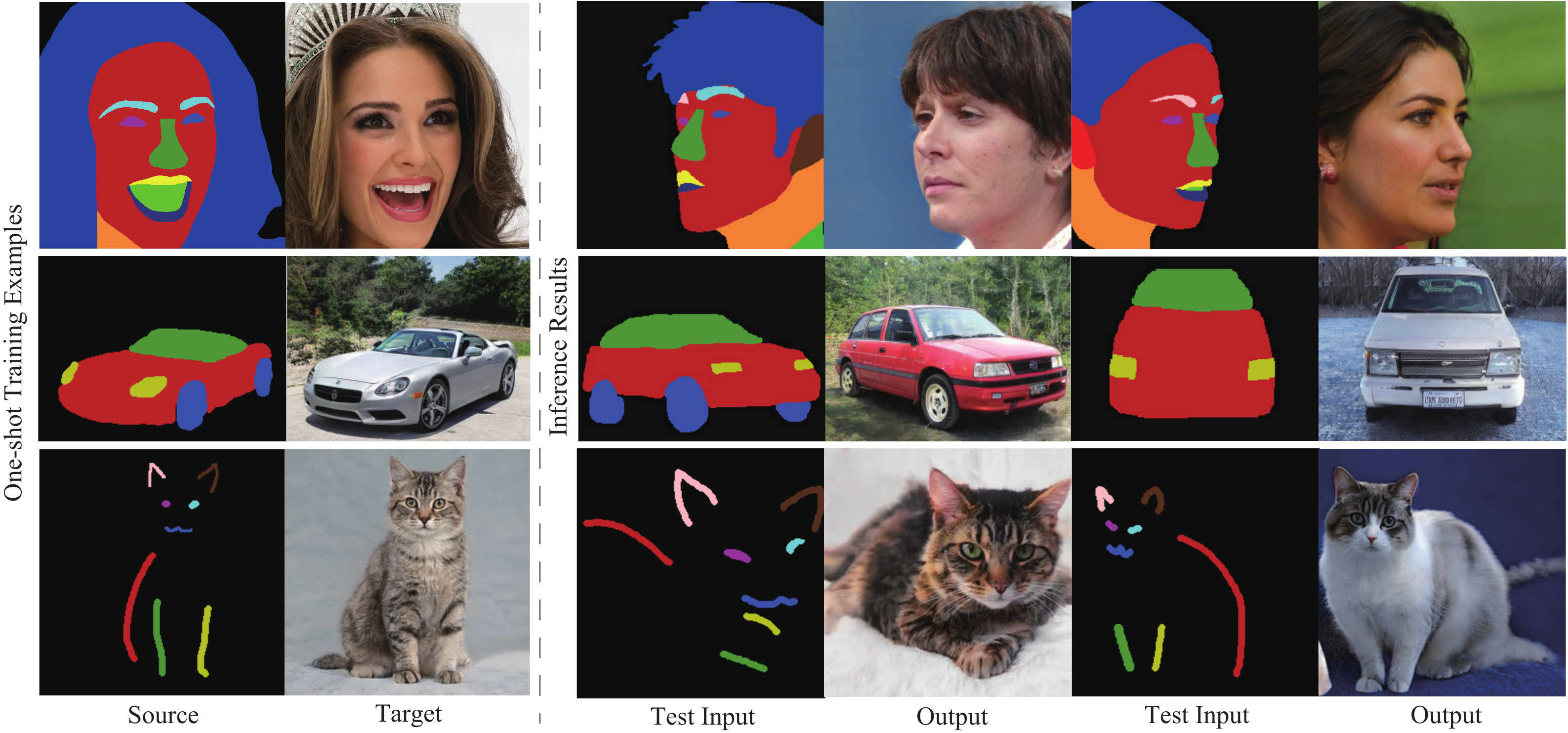}
    \captionof{figure}{
    Our method can synthesize photorealistic images from dense or sparse semantic annotations using a single training pair and a pre-trained StyleGAN. The source codes are available at~\href{https://github.com/endo-yuki-t/Fewshot-SMIS}{\chkURL{https://github.com/endo-yuki-t/Fewshot-SMIS}}.
    }
    \label{fig:teaser}
\end{center}%
}]

\maketitle

\begin{abstract}
   This paper tackles a \chkA{challenging} problem of generating photorealistic images from semantic layouts in few-shot scenarios 
   where annotated training pairs are hardly available \chkA{but} pixel-wise annotation is quite costly.
   \chkA{We present a training strategy that performs pseudo labeling of semantic masks using the StyleGAN prior. 
   Our key idea is to construct a simple mapping between the StyleGAN feature and each semantic class from a few examples of semantic masks. 
   With such mappings, we can generate an unlimited number of pseudo semantic masks from random noise to train an encoder for controlling a pre-trained StyleGAN generator.
   Although the pseudo semantic masks might be too coarse for previous approaches that require pixel-aligned masks, our framework can synthesize high-quality images from not only dense semantic masks but also sparse inputs such as landmarks and scribbles.}
   Qualitative and quantitative results \chkA{with} various datasets demonstrate improvement over previous approaches with respect to layout fidelity and visual quality in \chkA{as few as one- or five-shot settings}.
\end{abstract}

\section{Introduction}
Semantic image synthesis is a powerful technique for generating images with intuitive control using spatial semantic layouts. 
A drawback is that \chkG{most} existing techniques require substantial training data in source and target domains for high-quality outputs. 
Even worse, annotations of pixel-wise labels (e.g., semantic masks) are quite costly.

In this paper, we present the first method for few-shot semantic image synthesis, assuming that we can utilize many unlabeled data with only a few labeled data of the target domain.
Imagine that you have a large dataset of \chkA{car or} cat photos, but only a single annotated pair is available (Figure~\ref{fig:teaser}, \chkG{the} \chkA{2nd and 3rd rows}). 
In this scenario, we utilize the state-of-the-art generative adversarial network (GAN), \textit{StyleGAN}~\cite{DBLP:conf/cvpr/KarrasLA19,DBLP:conf/cvpr/KarrasLAHLA20}, \chkG{pre-trained using the unlabeled dataset.}
Namely, we \chkG{achieve high-quality image synthesis by exploring StyleGAN's latent space via \textit{GAN inversion}.}
\chkG{What is challenging here is that, although common GAN inversion techniques~\cite{DBLP:conf/iccv/AbdalQW19, DBLP:conf/cvpr/AbdalQW20} assume that test inputs belong to the same domain as GAN's training data (\eg, facial photographs), our test and training data are in different domains, \ie, semantic layouts and photographs. 
How to invert the input in a different domain into GAN's latent space is an open question, as introduced in the latest survey~\cite{xia2021gan}.}

\chkG{To bridge the domain gaps for the first time, we construct a mapping between the semantics predefined in the few-shot examples and StyleGAN's latent space.}
\chkG{Inspired by the fact that pixels with the same semantics tend to have similar StyleGAN features~\cite{DBLP:conf/cvpr/CollinsBPS20}, we generate pseudo semantic masks from random noise in StyleGAN's latent space via simple nearest-neighbor matching.}
This \chkG{way, we can} draw an unlimited number of training pairs by only feeding random noise to the \chkA{pre-trained} StyleGAN generator.
After integrating an encoder on top of the fixed StyleGAN generator, we then train the encoder for controlling the generator \chkC{using the pseudo-labeled data} in a supervised fashion. %
\chkC{Although our pseudo semantic masks might be too noisy or coarse for the previous pixel-aligned approach~\cite{DBLP:conf/cvpr/Park0WZ19}, our method works well with such masks thanks to the tolerance to misalignment.}
\chkF{\chkG{Our approach integrates} semantic layout control \chkG{into} pre-trained StyleGAN models publicly available on the Web~\cite{awesomeStyleGAN2}, \chkG{via pseudo labeling} even from a single annotated pair with \chkC{not only a dense mask but also} sparse scribbles or landmarks.}

\chkF{
In summary, our major contributions are three-fold:
\begin{itemize}
\item We explore a novel problem of \chkG{few-shot} semantic image synthesis, where the users can synthesize \chkG{high-quality, various} images in \chkG{the target domains} even from \chkG{very few and} rough semantic \chkG{layouts provided during training}. 
\item We propose a simple yet effective method for training a StyleGAN encoder for semantic image synthesis in \chkG{few-shot} scenarios, via pseudo sampling and labeleing based on \chkG{the} StyleGAN prior, without hyper parameter tuning for \chkG{complicated} loss functions. 
\item \chkG{We} demonstrate that our method significantly outperforms the existing methods w.r.t. layout fidelity and visual quality \chkG{via extensive experiments on various datasets.} 
\end{itemize}
}

\section{Related Work}
\paragraph{Image-to-Image translation}
\chkF{T}here are various \chkG{image-to-image (I2I) translation} methods suitable for semantic image synthesis\chkG{; the goals are, \eg,} to improve image quality~\cite{DBLP:conf/iccv/ChenK17,DBLP:conf/nips/LiuYS0L19,DBLP:conf/cvpr/Park0WZ19,DBLP:conf/cvpr/000500TS20,DBLP:conf/mm/0005BS20}, generate multi-modal outputs~\cite{DBLP:conf/cvpr/Park0WZ19,DBLP:conf/iccv/LiZM19,DBLP:conf/cvpr/ZhuXYB20,DBLP:journals/cgf/EndoK20}, and simplify input annotations using bounding boxes~\cite{DBLP:conf/cvpr/ZhaoMYS19,DBLP:conf/iccv/SunW19,DBLP:conf/cvpr/LiCGYWL20}. 
However, all of these methods require \chkA{large \chkG{amounts}} of training data of both source and target domains and thus are \chkA{unsuitable} for our \chkA{few-shot scenarios}.

\begin{figure}[t]
  \centering
  \includegraphics*[width=0.9\linewidth, clip]{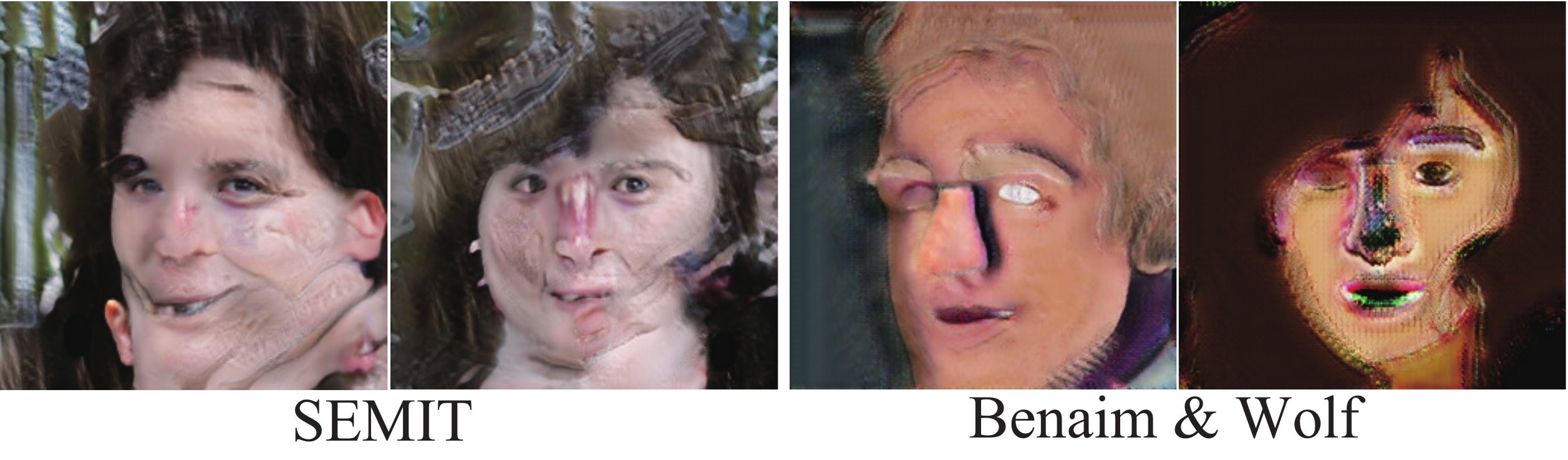}
  \caption{\chkF{%
  Results of \chkG{one-shot} semantic image synthesis with general few-shot I2I translation (SEMIT~\cite{DBLP:conf/cvpr/WangKG0K20} and Benaim and Wolf~\cite{DBLP:conf/nips/BenaimW18}) using the same training data as ours. The input semantic masks are the same as \chkG{those used} in Figure~\ref{fig:face}. 
  }
}
  \label{fig:semit}
\end{figure}

\chkA{FUNIT~\cite{DBLP:conf/iccv/0001HMKALK19} and SEMIT~\cite{DBLP:conf/cvpr/WangKG0K20} are recently proposed methods for \chkG{``\textit{few-shot}'' I2I} translation among different classes of photographs (\eg, dog, bird, and flower).
\chkG{However, their meaning of ``few-shot'' is quite different from ours;
they mean} that only a few target class data are available in test time\chkG{, but}
assume sufficient data of both source and target classes in training time (with a difference in whether the image class labels are fully available~\cite{DBLP:conf/iccv/0001HMKALK19} or not~\cite{DBLP:conf/cvpr/WangKG0K20}).}
Contrarily, we assume only a few source data, \ie, ground-truth \chkD{(GT)} semantic masks, in training time.
\chkG{These ``few-shot'' I2I translation methods do not work at all in our settings,} \chkA{as shown in Figure~\ref{fig:semit}.}

\chkF{
Benaim and Wolf~\cite{DBLP:conf/nips/BenaimW18} presented \chkG{a one-shot} unsupervised I2I translation framework for the same situation as ours. 
However, their ``\textit{unpaired}'' approach suffers from handling semantic \chkG{masks, which} have less distinctive features than \chkG{photographs} (see Figure~\ref{fig:semit}). 
Moreover, \chkG{their trained model has low generalizability, specialized for the single source image provided during training.} 
In other words, their method needs to train a model for each test input, while our method does not. 
Table~\ref{tab:precondition} summerizes \chkG{the} difference\chkG{s} of problem settings between each method. 
}

\begin{table}[t]
\caption{\chkF{%
\chkG{Feeding types and required amounts} of data \chkG{for} existing semantic image synthesis (SMIS), ``\textit{few-shot}`` image-to-image translation (I2I), and ours. 
}}
\small
\begin{tabular}{l|c|c|c|c}
\multicolumn{1}{c|}{}                                                          & \multicolumn{1}{l|}{}         & \multicolumn{2}{c|}{Training}                                         & \multicolumn{1}{l}{Test}   \\ \cline{3-5} 
Method                                                                         &
\multicolumn{1}{l|}{\chkG{Feeding}} &
\multicolumn{1}{l|}{source} & \multicolumn{1}{l|}{target} & \multicolumn{1}{l}{target} \\ \hline \hline
\chkG{SMIS \cite{DBLP:conf/cvpr/IsolaZZE17,DBLP:conf/cvpr/Park0WZ19,DBLP:conf/nips/LiuYS0L19}} & paired                        & large                                   & large                       & none                       \\ \hline
\begin{tabular}[c]{@{}l@{}}``Few-shot'' I2I~\cite{DBLP:conf/iccv/0001HMKALK19,DBLP:conf/cvpr/WangKG0K20}\\
\end{tabular}          & unpaired                      & large                                   & large                       & small                      
\\ \hline
Benaim and Wolf~\cite{DBLP:conf/nips/BenaimW18}                                                                         & unpaired                        & small                                   & large                       & none    
\\ \hline
 \chkG{Few-shot SMIS (ours)}                                                                           & paired                        & small                                   & large                       & none                      
\end{tabular}
\label{tab:precondition}
\end{table}

\paragraph{Latent space manipulation}

\chkF{\chkG{R}ecent GAN inversion (e.g., Image2StyleGAN~\cite{DBLP:conf/iccv/AbdalQW19, DBLP:conf/cvpr/AbdalQW20}) can control GAN outputs by inverting given images into \chkG{GAN's} latent space.
\chkF{There have been also many attempts to manipulate inverted codes in disentangled latent spaces~\cite{DBLP:journals/tog/ChiuKLIY20,gansteerability,DBLP:conf/cvpr/ShenGTZ20,DBLP:journals/corr/abs-2007-06600,DBLP:journals/corr/abs-2004-02546}.}
However, \chkG{inverting semantic masks into a latent space defined by photographs is not straightforward because how to measure the discrepancy between the two different domains (\ie, semantic masks and photographs) is an open question~\cite{xia2021gan}.}
\chkG{Note that} we cannot use pre-trained segmentation networks in our few-shot scenrarios.}
\chkG{Our method is the first attempt of GAN inversion for semantic masks into StyleGAN's latent space defined by photographs.}

\paragraph{Few-shot semantic image synthesis}
To the best of our knowledge, there is no \chkA{other} \chkF{few-shot method dedicated to} semantic image synthesis.
An alternative approach might be to use few-shot semantic  segmentation~\cite{DBLP:conf/bmvc/DongX18,DBLP:conf/iccv/WangLZZF19,DBLP:conf/mm/LiuCLGCT20,DBLP:conf/aaai/TianWQWSG20,WangECCV20,YangECCV20} to annotate unlabeled images to train image-to-image translation models.
In recent few-shot semantic segmentation methods based on a meta-learning approach, however, \chkD{training episodes require large numbers of labeled images of various classes other than target classes} to obtain common knowledge. Therefore, this approach is not applicable to our problem setting. 

\section{Few-shot Semantic Image Synthesis}
\subsection{Problem setting}
\label{sec:ProblemSetting}
Our goal is to accomplish semantic image synthesis via semi-supervised \chkA{learning} with $N_u$ unlabeled images and $N_l$ labeled pairs both in the same target domain, where $N_u \gg N_l$.
In particular, we assume few-shot scenarios, setting $N_l = 1$ or $5$ in our results.
A labeled pair consists of a one-hot semantic mask $\mathbf{x}\in\{0,1\}^{C\times W\times H}$ (where $C$, $W$, and $H$ are the number of classes, width, and height) and its \chkD{GT} RGB image $\mathbf{y}\in\mathbb{R}^{3\times W\times H}$. 
A semantic mask can be a dense map pixel-aligned to $\mathbf{y}$ or a sparse map (e.g., scribbles or landmarks).
In a sparse map, each scribble or landmark has a unique class label, whereas unoccupied pixels have an ``\textit{unknown}'' class label.
Hereafter we denote the labeled dataset as $\mathcal{D}_l = \{\mathbf{x}_i,\mathbf{y}_i\}^{N_l}_{i=1}$ and the unlabeled dataset as $\mathcal{D}_u=\{\mathbf{y}_i\}^{N_u}_{i=1}$.

\subsection{Overview}

The core of our method is \chkG{to find appropriate mappings between semantics defined by a few labeled pairs $\mathcal{D}_l$ and StyleGAN's latent space defined by} an unlabeled dataset $\mathcal{D}_u$. 
\chkG{Specifically,} we first extract a feature vector representing each semantic class, which we refer to as a \textit{representative vector}, \chkG{and then find matchings with StyleGAN's feature map via ($k$-)nearest-neighbor search. 
Such matchings enable \textit{pseudo labeling}, \ie, to obtain pseudo semantic masks from random noise in StyleGAN's latent space, which are then used to train an encoder for controlling the pre-trained StyleGAN generator.} 
A similar approach is the prototyping used in recent few-shot semantic segmentation~\cite{DBLP:conf/bmvc/DongX18,DBLP:conf/iccv/WangLZZF19,YangECCV20}.
\chkD{Our advantage is that our method suffices with unsupervised training of StyleGAN models, whereas the prototyping requires supervised training of feature extractors (e.g., VGG~\cite{DBLP:journals/corr/SimonyanZ14a}).}

\begin{figure}[t]
  \centering
  \includegraphics*[width=1\linewidth, clip]{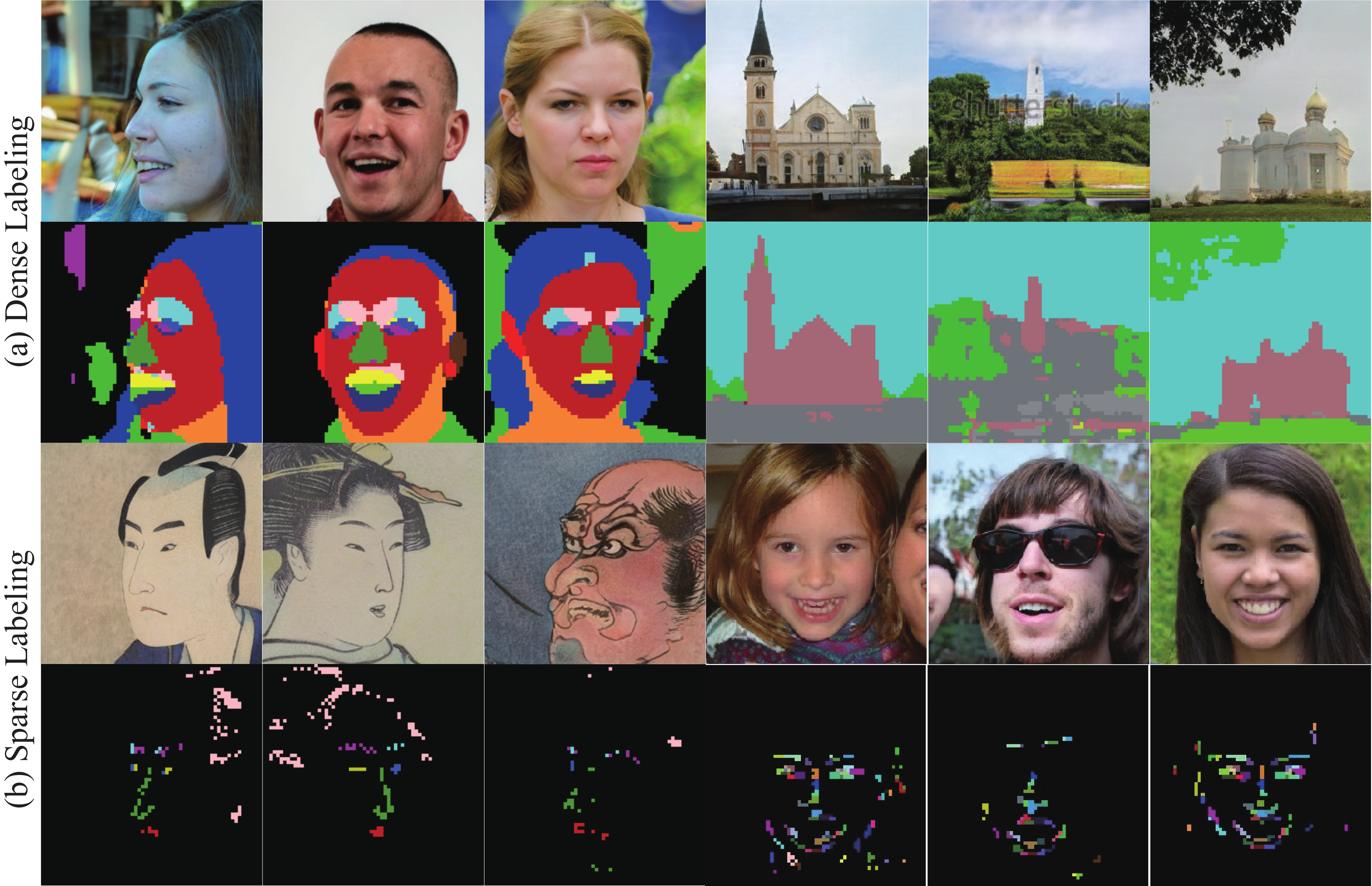}
  \caption{
  Our densely (a) and sparsely (b) pseudo-labeled examples. 
}
  \label{fig:pseudo_sample}
\end{figure}

Our pseudo semantic masks \chkA{are} often \chkA{noisy, distorted \chkC{(see Figure~\ref{fig:pseudo_sample})},} and thus inadequate for conventional approaches of semantic image synthesis or image-to-image translation, which require pixel-wise correspondence.
However, even from such low-quality pseudo semantic masks, we can synthesize high-quality images with spatial layout control by utilizing the \chkA{pre-trained} StyleGAN generator.
This is because the StyleGAN generator only requires latent codes that encode spatially global information.

As an encoder for generating such latent codes, we adopt the \textit{Pixel2Style2Pixel} (pSp) encoder~\cite{DBLP:journals/corr/abs-2008-00951}.
The inference process is the same as that of pSp; from a semantic mask, the encoder generates latent codes that are then fed to the fixed StyleGAN generator to control the spatial layout.
We can optionally change or fix latent codes that control local details of the output images.
Please refer to Figure 3 in the pSp paper~\cite{DBLP:journals/corr/abs-2008-00951} for more details.

Hereafter we \chkA{explain} the pseudo labeling process and the training procedure with the pseudo semantic masks.

\subsection{Pseudo labeling}

\chkA{We} elaborate on how to calculate the representative vectors and pseudo labeling, for which we propose different approaches to dense and sparse semantic masks.

\subsubsection{Dense pseudo labeling}
\label{sec:DensePseudoLabeling}

\begin{figure}[t]
  \centering
  \includegraphics*[width=1.\linewidth, clip]{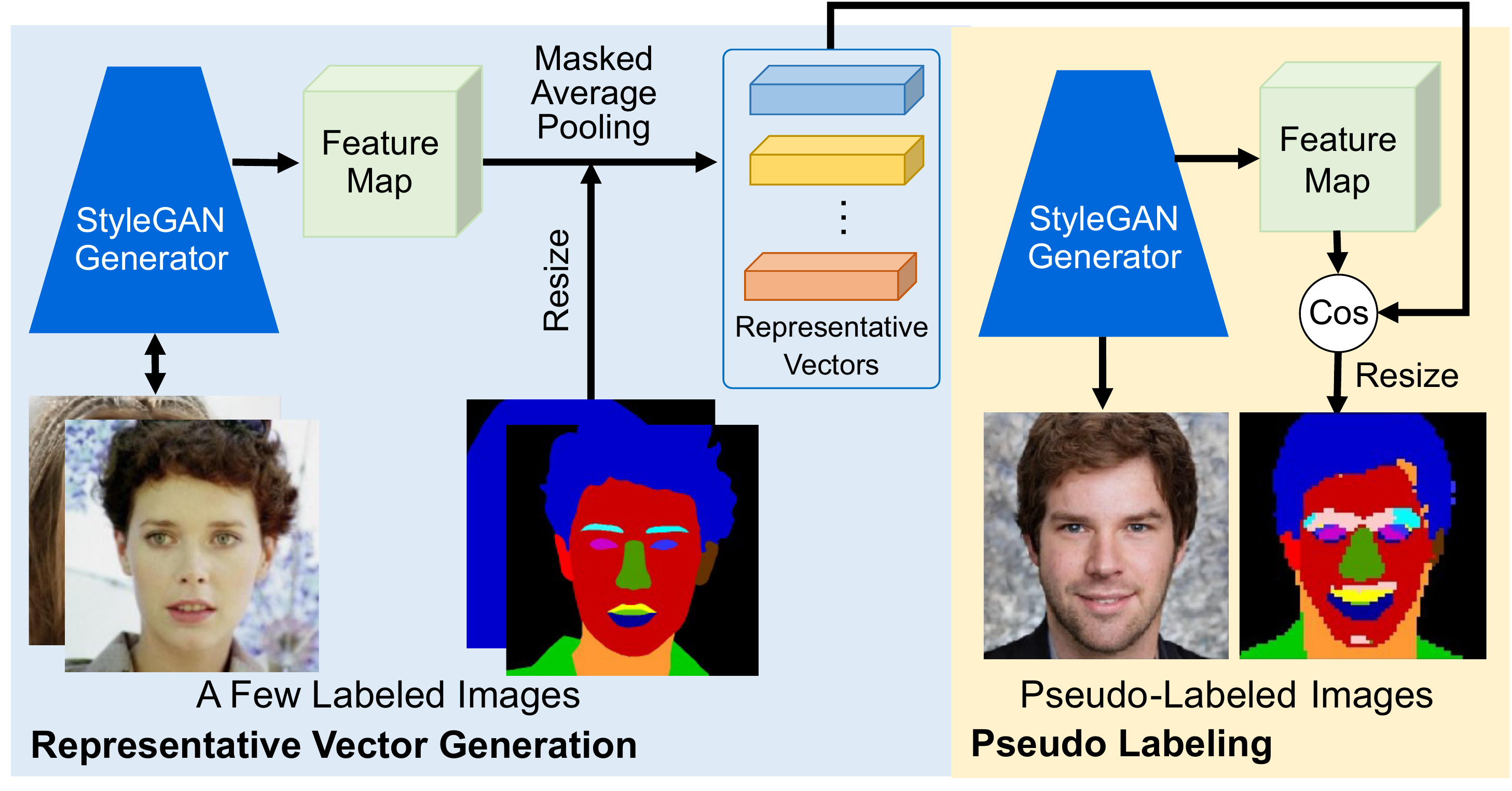}
  \caption{
  Dense pseudo labeling. 
  \chkA{Left:} We compute \chkA{a} representative vector of each class \chkA{via} masked average pooling \chkA{over} the feature maps of few-shot examples. 
  \chkA{Right:} We assign pseudo labels to sampled images \chkA{via} nearest-neighbor matching based on cosine similarity between the representative vectors and the feature maps of the sampled images. 
}
  \label{fig:dense}
\end{figure}
Figure~\ref{fig:dense} illustrates the pseudo labeling process for dense semantic masks. 
We first extract StyleGAN's feature maps corresponding to the semantic masks in $\mathcal{D}_l$. 
If pairs of semantic masks $\mathbf{x}$ and \chkD{GT} RGB images $\mathbf{y}$ are available in $\mathcal{D}_l$, we first invert $\mathbf{y}$ into the StyleGAN's latent space via optimization and then extract the feature map via forward propagation. 
Otherwise, we feed one or a few \chkA{noise vectors} to the \chkA{pre-trained} StyleGAN generator, extract the feature maps and synthesized images, and manually annotate the synthesized images to create semantic masks.
Next, we extract a representative vector $\mathbf{v}_c$ for each semantic class $c$ from the pairs of extracted feature maps and semantic masks, following the approach by Wang~\etal~\cite{DBLP:conf/iccv/WangLZZF19} for prototyping.
Specifically, we apply the masked average pooling to the feature map $\mathbf{F}_i \in \mathbb{R}^{Z\times W' \times H'}$ (where $Z$, $W'$, and $H'$ are the number of channels, width, and height) using a resized semantic mask $\mathbf{x}'_i \in \mathbb{R}^{C\times W' \times H'}$, and then average over each pair $i$ in $\mathcal{D}_l$:
\begin{align}
\mathbf{v}_c = \frac{1}{N_l}\sum_{i=1}^{N_l}\frac{\sum_{x,y}\mathbf{F}^{x,y}_i\1\left[\mathbf{x}^{(c,x,y)}_i=1 \right]}{\sum_{x,y}\1\left[\mathbf{x}^{(c,x,y)}_i=1 \right]},
\end{align}
where $(x, y)$ denote pixel positions, and $\1 \left[\cdot \right]$ is the indicator function that returns 1 if the argument is true and 0 otherwise.

After obtaining representative vectors, we generate pseudo semantic masks for training our encoder.
Every time we feed \chkA{random noise} to the \chkA{pre-trained} StyleGAN generator, we extract a feature map $\mathbf{F'}$ and then calculate a semantic mask via nearest-neighbor matching between the representative vectors and the pixel-wise vectors in $\mathbf{F'}$.
\chkA{In all of our results, feature maps $\mathbf{F'}$ are at resolution of $64 \times 64$ and extracted from the layer closest to the output layer of the StyleGAN generator.}
Class label $c^{(x,y)}$ for pixel $(x, y)$ is calculated as follows:
\begin{align}
c^{(x,y)} = \mathrm{argmax}_{c\in C} \cos(\mathbf{v}_c, \mathbf{F'}^{(x,y)}). 
\end{align}
As a distance metric, we adopt the cosine similarity $\cos(\cdot,\cdot)$, inspired by the finding~\cite{DBLP:conf/cvpr/CollinsBPS20} that StyleGAN's feature vectors \chkA{having} the same semantics form clusters on a unit sphere.
Finally, we enlarge the semantic masks to the size of the synthesized images.
Figure~\ref{fig:pseudo_sample}(a) shows the examples of pseudo labels for dense semantic masks.

\subsubsection{Sparse pseudo labeling}
\begin{figure}[t]
  \centering
  \includegraphics*[width=1.\linewidth, clip]{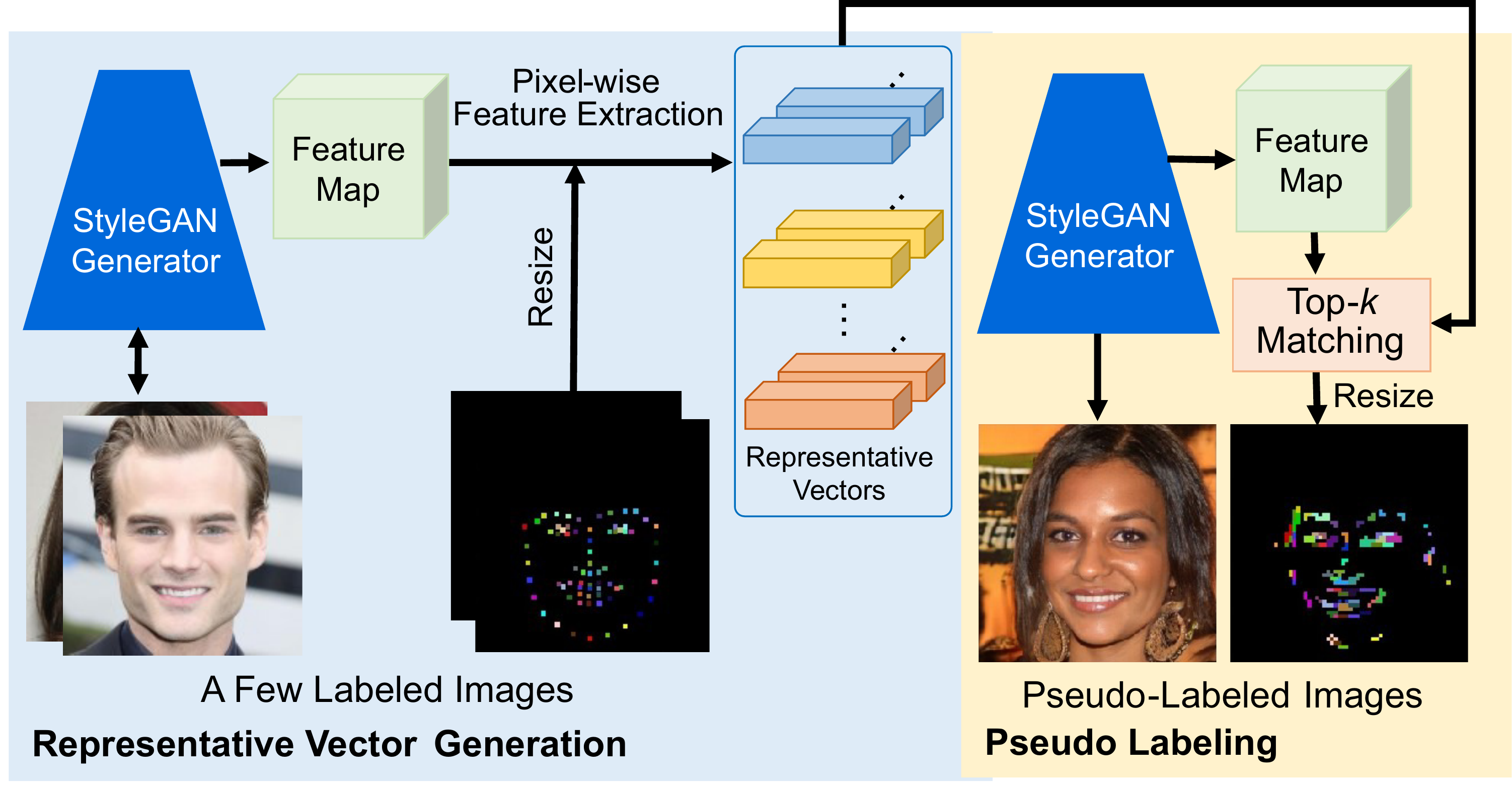}
  \caption{
  Sparse pseudo labeling. 
  \chkA{Left:} Unlike the dense \chkA{version}, we extract representative vectors for all labeled pixels.
  \chkA{Right:} For each representative vector, we take top-$k$ correspondences and assign its class label to the corresponding pixels \chkA{whose similarites are} above a threshold $t$. 
}
  \label{fig:sparse}
\end{figure}

Figure~\ref{fig:sparse} illustrates the pseudo labeling process for sparse semantic masks. 
As explained in Subsection~\ref{sec:ProblemSetting}, sparse semantic masks have a class label for each annotation (e.g., a scribble and landmark) and an ``unknown'' label. 
Here we adopt a pseudo-labeling approach different from the dense version due to the following reason.
We want to retain the spatial sparsity in pseudo semantic masks so that the pseudo semantic masks resemble genuine ones as much as possible.
However, if we calculate nearest-neighbors for a representative vector of each annotation as done in the dense version, the resultant pseudo masks might form dense clusters of semantic labels.
\chkG{Alternatively, as a simple heuristics, we consider each pixel in each annotation has its representative vector and calculate a one-to-one matching between each annotated pixel and each pixel-wise vector. 
In this case, however, many annotated pixels might match an identical pixel-wise vector (\ie, many-to-one mapping),} which results in fewer samples in pseudo semantic masks. 
Therefore, we calculate \chkA{top-$k$ (i.e., }$k$-nearest-neighbors\chkA{)} instead of one-nearest-neighbor to increase matchings.
In the case of many-to-one mappings, we assign the class label of an annotation that has the largest cosine similarity.
\chkG{To avoid outliers, we discard the matchings} if their cosine similarities are \chkG{lower than} a threshold $t$ \chkG{and} assign the ``unknown'' label.
Figure~\ref{fig:pseudo_sample}(b) shows the examples of pseudo labels for sparse semantic masks.

\chkA{We set $k=3$ and $t=0.5$ in all of our results in this paper.
The supplementary material contains pseudo-labeled results with different parameters.}

\subsection{Training procedure}
\begin{figure}[t]
  \centering
  \includegraphics*[width=1.\linewidth, clip]{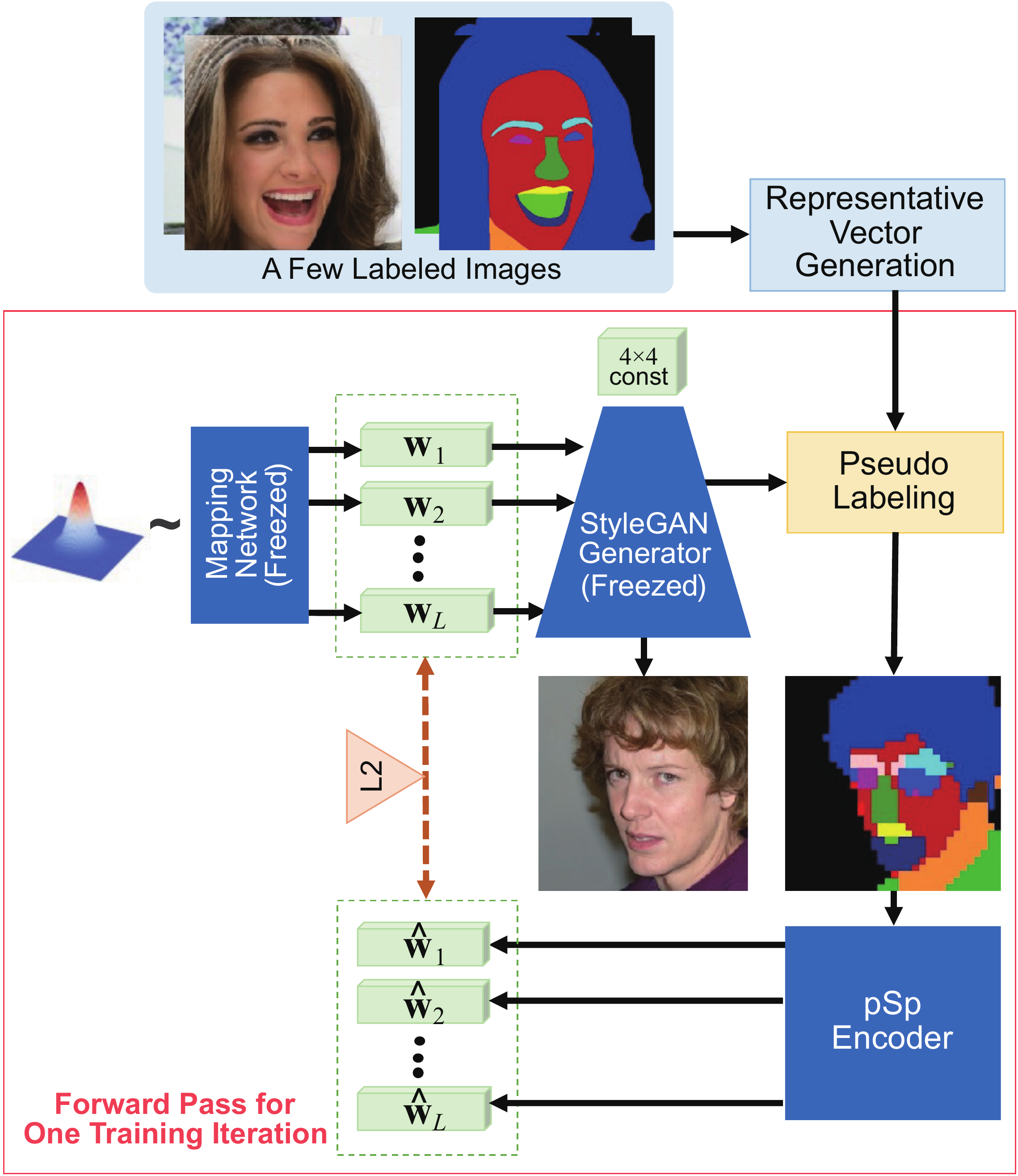}
  \caption{%
  Training \chkA{iteration} of the encoder.
  \chkA{We first} generate images from noise vectors via the mapping network and the StyleGAN generator. 
  We then compute pseudo semantic masks using the \textit{representative vectors} \chkA{(Figures~\ref{fig:dense} and \ref{fig:sparse})}.  
    \chkA{We} optimize the encoder parameters \chkA{based on L2 loss between latent codes.}
}
  \label{fig:training}
\end{figure}

Figure~\ref{fig:training} illustrates the learning process of our encoder.
First, we explain the forward pass in the training phase.
We feed a random noise $\mathbf{z}$ sampled from a normal distribution $\mathcal{N}(\mathbf{0},\mathbf{I})$ to the encoder and obtain latent codes $\{\mathbf{w}_i\}^L_{i=1}$ (where $L$ is the number of layers to input/output latent codes) via the \chkA{pre-trained StyleGAN's} mapping network $f$.
We feed the latent codes to the \chkA{pre-trained} StyleGAN generator to synthesize an image while extracting the intermediate layer's feature map.
From this feature map and representative vectors, we create a pseudo semantic mask, which is then fed to our encoder to extract latent codes $\{\mathbf{\hat{w}}_i\}^L_{i=1}$.

In the backward pass, we optimize the encoder using the following loss function:
\begin{align}
\mathcal{L} = \mathbb{E}_{\mathbf{w}\sim f(\mathbf{z})}{\|\mathbf{\hat{w}}-\mathbf{w}\|^2_2}. 
\label{eq:loss}
\end{align}
This loss function indicates that our training is quite simple because backpropagation does not go through the \chkA{pre-trained} StyleGAN generator.
Algorithm~\ref{alg1} summarizes the whole process of training.
In the supplementary material, we also show the intermediate pseudo semantic masks and reconstructed images obtained during the training iterations.

\begin{algorithm}                      
\caption{Few-shot learning of StyleGAN encoder}         
\label{alg1}                          
\begin{algorithmic}                  
\REQUIRE A labeled set $\mathcal{D}_l$ and unlabeled set $\mathcal{D}_u$
\STATE Train StyleGAN using $\mathcal{D}_u$
\STATE Compute representative vectors using $\mathcal{D}_l$
\FOR {each training iteration}
\STATE Sample latent codes according to $\mathcal{N}(\mathbf{0}, \mathbf{I})$
\STATE Feed the latent codes to the generator
\STATE Pseudo labeling using representative vectors
\STATE Feed the pseudo semantic masks to the encoder
\STATE Compute the loss $\mathcal{L}$ as in \chkD{Eq.~(\ref{eq:loss})}
\STATE Compute the gradient and optimize the encoder
\ENDFOR
\end{algorithmic}
\end{algorithm}

\section{Experiments}
\begin{figure*}[t]
  \centering
  \includegraphics*[width=1.\linewidth, clip]{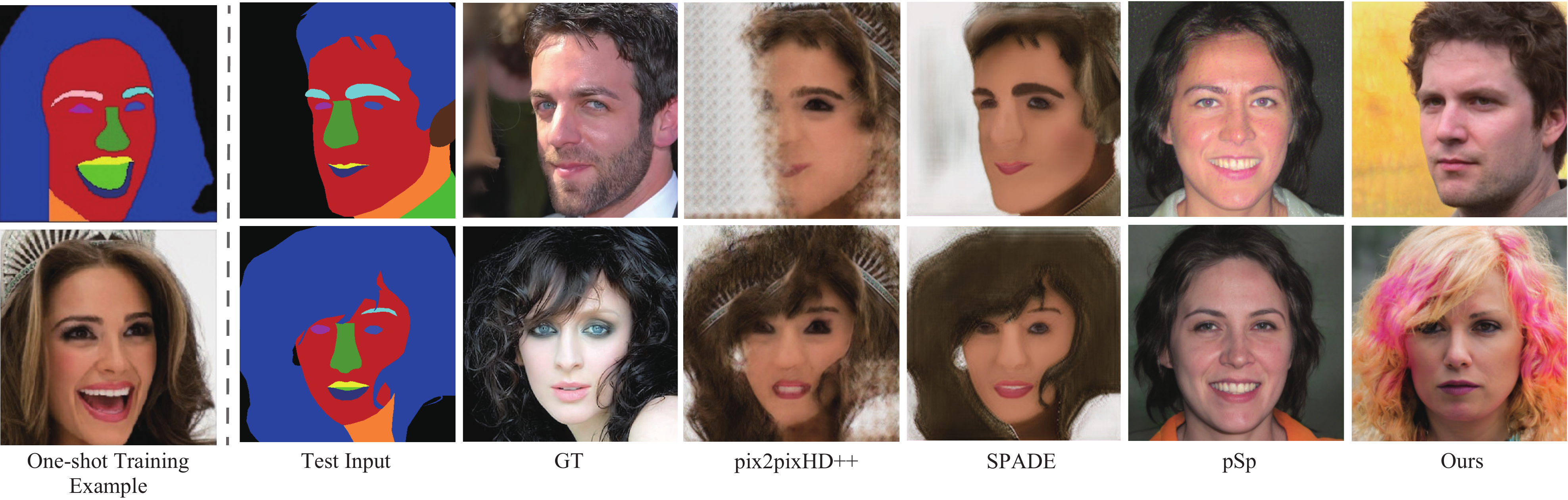}
  \caption{%
  Comparison of face images generated from dense semantic masks in a one-shot setting. 
}
  \label{fig:face}
\end{figure*}

\begin{figure*}[t]
  \centering
  \includegraphics*[width=1.\linewidth, clip]{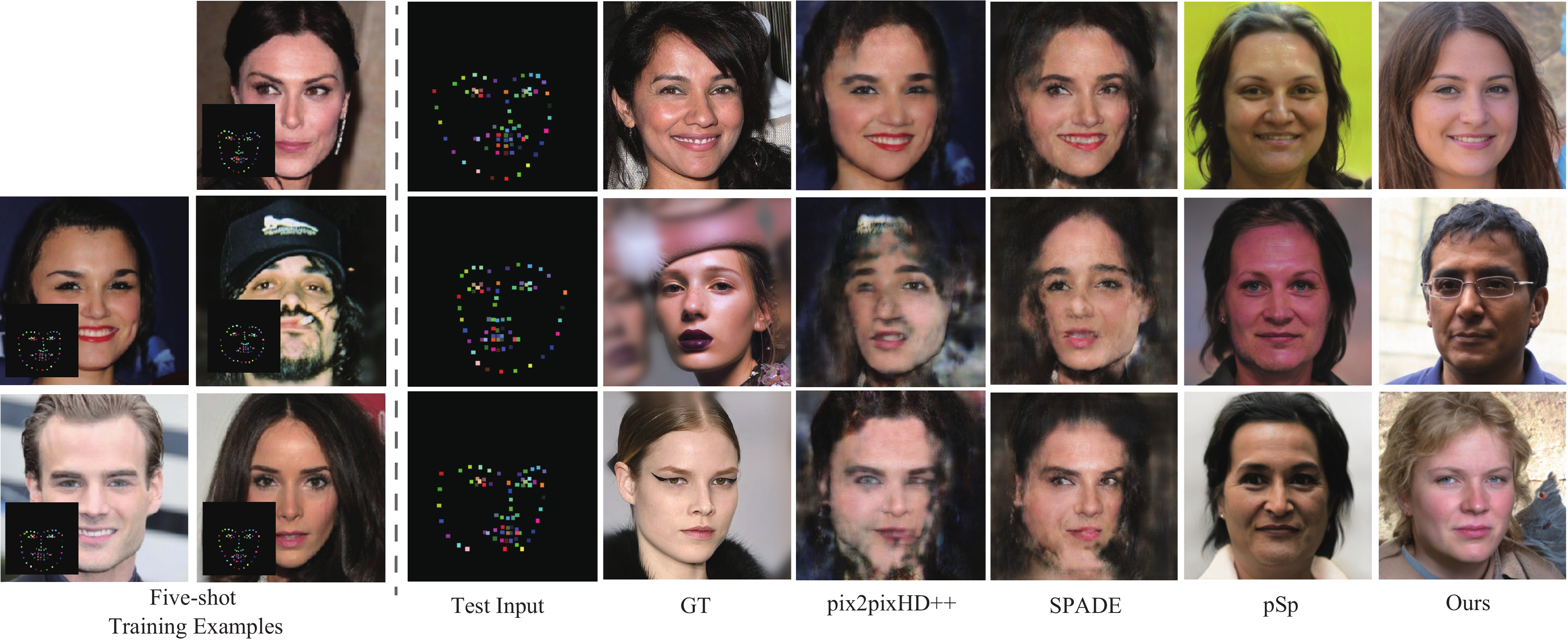}
  \caption{%
  Comparison of face images generated from sparse landmarks in a five-shot setting. 
}
  \label{fig:facelandmark}
\end{figure*}

\begin{figure*}[t]
  \centering
  \includegraphics*[width=1.\linewidth, clip]{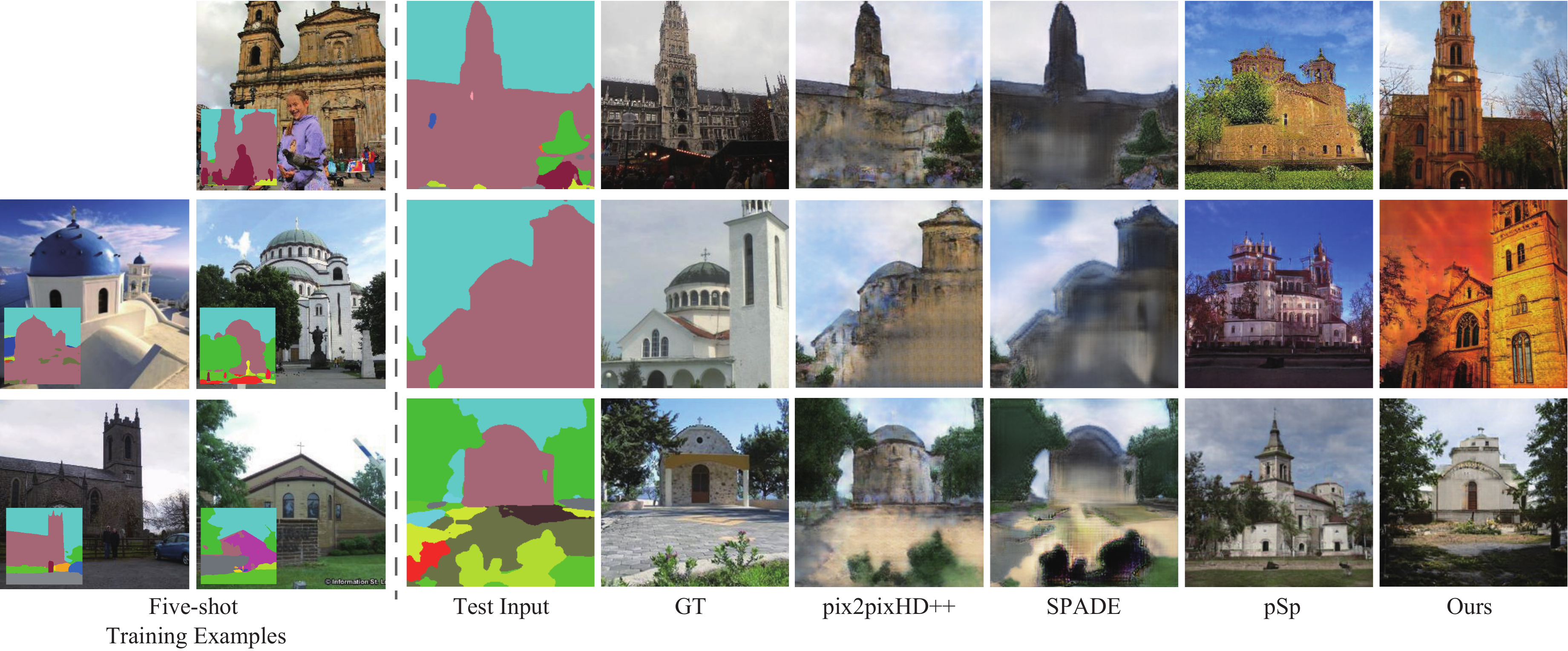}
  \caption{%
  Comparison of church images generated from dense semantic masks in a five-shot setting. 
}
  \label{fig:church}
\end{figure*}

\begin{figure}[t]
  \centering
  \includegraphics*[width=1.\linewidth, clip]{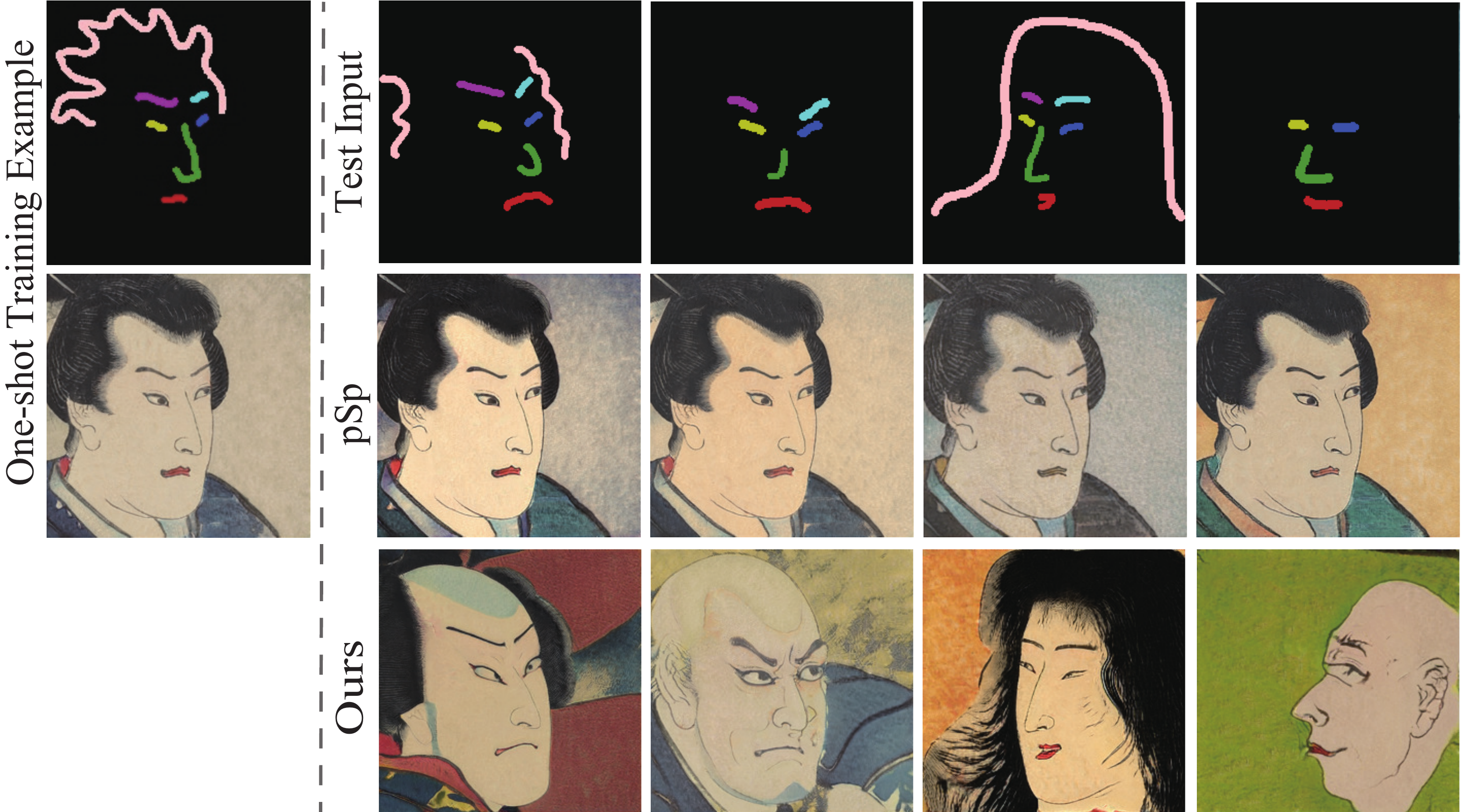}
  \caption{%
  Comparison of ukiyo-e images generated from \chkA{sparse scribbles} in a five-shot setting.
}
  \label{fig:ukiyoe}
\end{figure}

\begin{figure}[t]
  \centering
  \includegraphics*[width=1.\linewidth, clip]{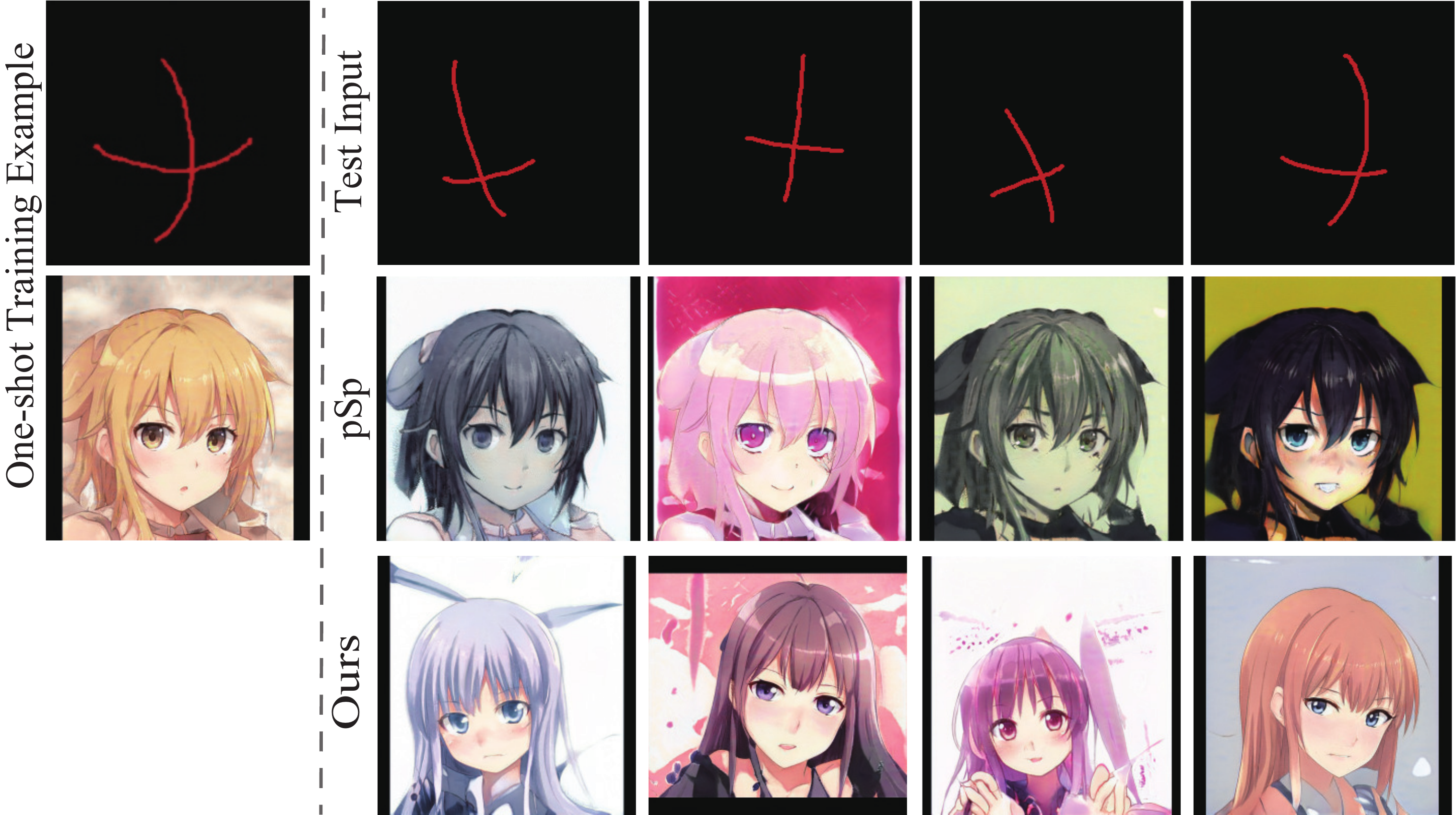}
  \caption{%
  Comparison of anime \chkD{face images} generated from \chkA{cross lines} in a five-shot setting. 
}
  \label{fig:manga}
\end{figure}

\begin{figure}[t]
  \centering
  \includegraphics*[width=1\linewidth, clip]{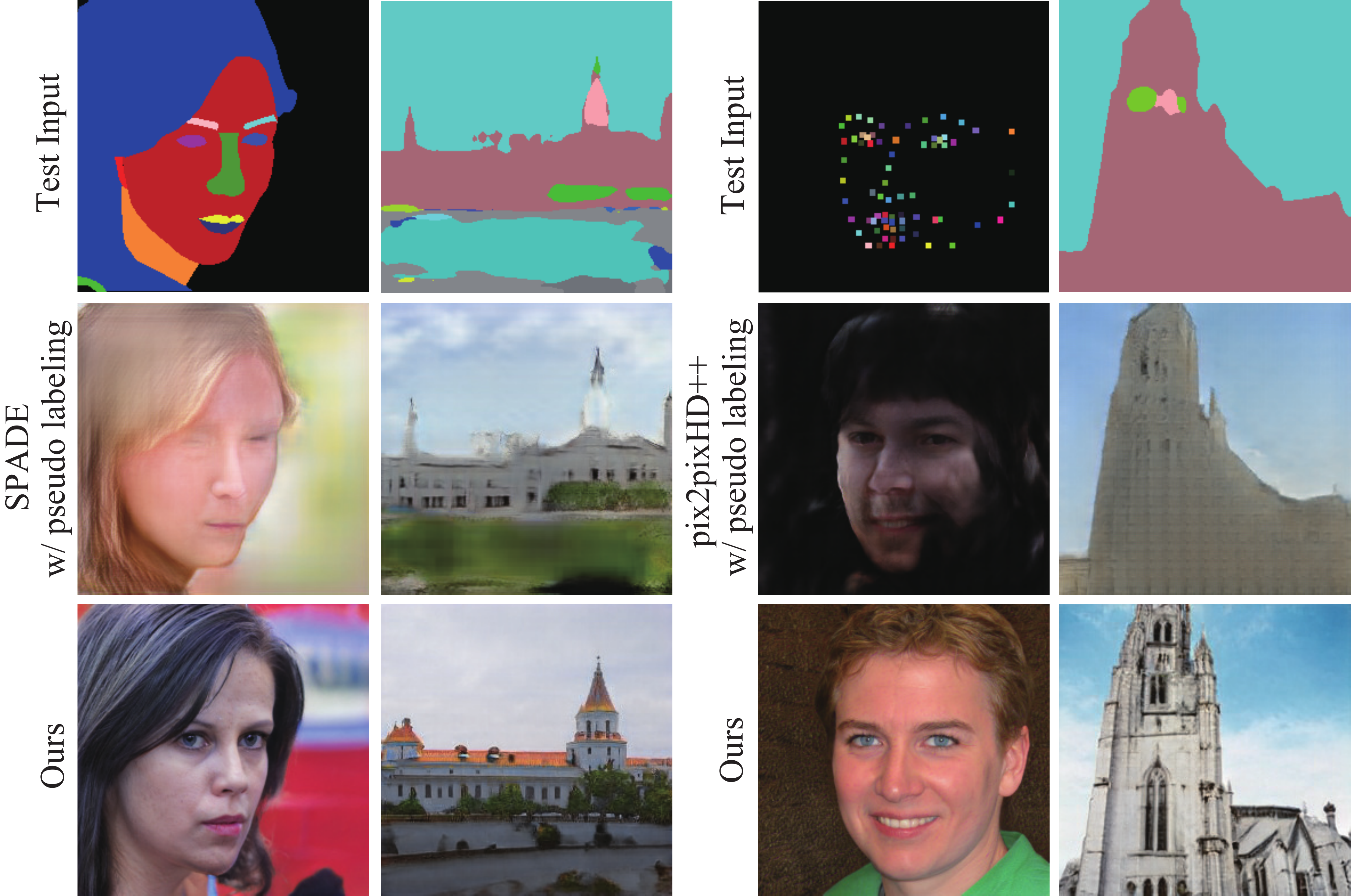}
  \caption{%
  Comparison with \chkA{the pixel-aligned} approach (SPADE~\cite{DBLP:conf/cvpr/Park0WZ19} and pix2pixHD++~\cite{DBLP:conf/cvpr/Park0WZ19}) trained with our pseudo-labeled data. 
}
  \label{fig:comparison_w_pseudolabels}
\end{figure}

\chkD{We conducted experiments to evaluate our method. The supplementary material contains implementation details. 
}

\subsection{Datasets}
\label{sec:Datasets}
We used public \chkD{StyleGAN2~\cite{DBLP:conf/cvpr/KarrasLAHLA20}} models pre-trained with FFHQ (human faces)~\cite{DBLP:conf/cvpr/KarrasLA19,DBLP:conf/cvpr/KarrasLAHLA20}, LSUN (car, cat, and church)~\cite{DBLP:journals/corr/YuZSSX15,DBLP:conf/cvpr/KarrasLAHLA20}, \chkA{ukiyo-e}~\cite{ukiyoePinkney}, and anime face images~\cite{animeGokaslan}. 
To evaluate our method quantitatively, we used the pre-processed CelebAMask-HQ datasets~\cite{DBLP:conf/cvpr/ZhuAQW20}, which contains face images and corresponding semantic masks \chkC{(namely, 2,000 for test and 28,000 for training)}. 
In addition, we extracted \chkA{face} landmarks as sparse annotations using OpenPose~\cite{8765346}. 
\chkC{The numbers of ``\textit{ground-truth}'' face landmarks are reduced to 1,993 for test and 27,927 for training because OpenPose sometimes failed.}
We used also LSUN church~\cite{DBLP:journals/corr/YuZSSX15} \chkC{(\chkF{300 in a validation set and 1,000 in a training set})} for the quantitative evaluation. 
Because this dataset does not contain semantic masks, we prepared them using the scene parsing model~\cite{DBLP:conf/cvpr/ZhouZPFB017} consisting of \chkD{the} ResNet101 encoder~\cite{DBLP:conf/cvpr/HeZRS16} and \chkD{the} UPerNet101 decoder~\cite{DBLP:conf/eccv/XiaoLZJS18}. 
\chkA{For our experiments of \chkB{$N_l$}-shot learning,} we selected \chkB{$N_l$} \chkF{paired} images from the training \chkF{sets, while the full-shot version uses all of them. }

\subsection{Qualitative results}

Figure~\ref{fig:face} compares the results generated from semantic masks of the CelebAMask-HQ dataset in a one-shot setting. 
Figure~\ref{fig:facelandmark} also shows the results generated from sparse landmarks in a five-shot setting. 
The \chkA{pixel-aligned} approach, 
SPADE~\cite{DBLP:conf/cvpr/Park0WZ19} and pix2pixHD++~\cite{DBLP:conf/cvpr/Park0WZ19}, generates images faithfully to the given layouts, but the visual quality is very low. 
Meanwhile, pSp~\cite{DBLP:journals/corr/abs-2008-00951}, which uses pre-trained StyleGANs, is able to generate realistic images. 
Although \chkA{pSp} can \chkA{optionally generate} multi-modal outputs, \chkA{it ignores the input layouts due to over-fitting to the too few training examples}. 
In contrast, our method produces photorealistic images corresponding to the given layouts. 
\chkA{We can see the same tendency in} comparison \chkA{with} the LSUN church dataset in Figure~\ref{fig:church}.

The benefit of \chkA{our} few-shot learning approach is not to need many labeled data. 
We therefore validate the applicability of our method to various domains where annotations are hardly available in public. 
Figures~\ref{fig:teaser} and~\ref{fig:ukiyoe} show car, cat, and ukiyo-e images generated from semantic masks and scribbles. Again, pSp does not reflect the input layouts on the results, whereas our method controls output semantics accordingly (e.g., \chkD{the cats' postures and the ukiyo-e hairstyles}). 
Interestingly, our method \chkA{works well with cross lines as inputs}, which \chkA{specify the orientations of anime faces (Figure~\ref{fig:manga})}.

Finally, we \chkA{conducted a comparison with the pixel-aligned approach using our pseudo labeling technique}.
Figure~\ref{fig:comparison_w_pseudolabels} shows the \chkA{results of} SPADE, pix2pixHD++, \chkA{and ours, which were trained up to 100,000 iterations with the appropriate loss functions.}
\chkA{Because our pseudo semantic masks are often misaligned, the pixel-aligned approach failed to learn photorealistic image synthesis, whereas ours succeeded.}
Please refer to the supplementary material for more qualitative results. 

\subsection{Quantitative results}

We quantitatively evaluated \chkC{competitive methods and ours} with respect to layout fidelity and visual quality.
\chkC{For each dataset, we first generate images from test data (i.e., semantic masks/landmarks in CelebA-HQ and semantic masks in LSUN church) using each method and then extract the corresponding semantic masks/landmarks for evaluation, as done in Subsection~\ref{sec:Datasets}.}
As evaluation \chkA{metrics} for parsing, we used Intersection over Union (IoU) and accuracy. 
As for IoU, we used mean IoU \chkD{(mIoU)} for CelebA-HQ.
\chkA{For LSUN church, we used frequency weighted IoU (fwIoU) because our ``ground-truth'' (GT) semantic masks synthesized by~\cite{DBLP:conf/cvpr/ZhouZPFB017} often contain small noisy-labeled regions, which} strongly affect mIoU.
As a landmark metric, we computed RMSE of Euclidean distances between landmarks of generated and GT images. 
If landmarks cannot be detected in generated images, we counted them as \textit{N/A}.
\chkE{We} used Fr\'{e}chet Inception Distance (FID) as a metric for visual quality. 

\begin{table*}[t]
\centering
\caption{Quantitative comparison on each dataset. $N_l$ is the number of labeled training data,  \textit{*} means training \chkA{the model} using our pseudo-labeled images sampled from the pre-trained StyleGANs, and \textit{Full} means using all labeled data in each traininig set. \textit{N/A} indicates the number of images in which landmarks cannot be detected. 
}
\begin{tabular}{r|c||c|c|c|c|c|c|ccc}
 \multicolumn{2}{c||}{} & \multicolumn{3}{c|}{CelebAMask-HQ} & \multicolumn{3}{c|}{CelebALandmark-HQ } & \multicolumn{3}{c}{LSUN ChurchMask}                \\ \cline{3-11} 
Method    & $N_l$       & mIoU$\uparrow$          & accu$\uparrow$          & FID$\downarrow$         & RMSE$\downarrow$         & N/A$\downarrow$        & FID$\downarrow$        & \multicolumn{1}{c|}{fwIoU$\uparrow$} & \multicolumn{1}{c|}{accu$\uparrow$} & FID$\downarrow$ \\ \hline \hline
pix2pixHD++~\cite{DBLP:conf/cvpr/Park0WZ19} & $5$ & \sbest{62.2}             & \sbest{92.7}             & 82.7           & 43.5           & 31           & 109.7          & \multicolumn{1}{c|}{71.8}    & \multicolumn{1}{c|}{78.3}    & 133.2   \\
pix2pixHD++* & 5 & 38.7             &  87.6             & 98.0           & \best{28.9}           & 40           & 125.6          & \multicolumn{1}{c|}{\sbest{68.0}}    & \multicolumn{1}{c|}{\sbest{76.7}}    & 100.6   \\
SPADE~\cite{DBLP:conf/cvpr/Park0WZ19} & 5 & \best{64.8}             & \best{92.4}             & 79.3           & \sbest{23.8}           & 35           & 121.4          & \multicolumn{1}{c|}{63.6}    & \multicolumn{1}{c|}{70.6}    & 193.2   \\
SPADE* & 5 & 40.1             & 88.8             & 88.5            & 33.7           & \sbest{11}           & 138.8        & \multicolumn{1}{c|}{\best{70.3}}    & \multicolumn{1}{c|}{\best{78.9}}    & 101.7   \\ \hline
pSp~\cite{DBLP:journals/corr/abs-2008-00951} & 1  & 24.8            & 61.7             & 93.4           & 37.0           & \best{0}           & 87.9          & \multicolumn{1}{c|}{29.2}    & \multicolumn{1}{c|}{42.0}    & 96.5   \\
pSp~\cite{DBLP:journals/corr/abs-2008-00951} & 5    & 28.4             & 67.6             & 96.9           & 37.1           & \best{0}           & 87.5          & \multicolumn{1}{c|}{34.4}    & \multicolumn{1}{c|}{47.9}    & 89.5   \\
Ours & 1  & 38.3             & 81.7             & \sbest{59.1}           & 31.5           & \best{0}           & \sbest{82.1}          & \multicolumn{1}{c|}{41.8}    & \multicolumn{1}{c|}{54.0}    & \sbest{65.1}   \\
Ours & 5  & 41.5             & 82.5             & \best{53.9}           & 31.4           & \best{0}           & \best{77.4}          & \multicolumn{1}{c|}{48.6}    & \multicolumn{1}{c|}{62.1}    & \best{56.9}   \\ \hline
pSp~\cite{DBLP:journals/corr/abs-2008-00951} & Full   & 49.2             & 87.9             & 67.4           & \chkF{26.6}          & \best{0}           & \chkF{106.6}          & \multicolumn{1}{c|}{50.4}    & \multicolumn{1}{c|}{63.8}    & 76.2  
\end{tabular}
\label{tab:eval}
\end{table*}

Table~\ref{tab:eval} shows the quantitative comparison \chkA{in few-shot settings, except for the bottom row, where} all labeled images in the training \chkA{datasets} were used.
\chkA{In the five-shot setting, the pixel-aligned approach (i.e., pix2pixHD++~\cite{DBLP:conf/cvpr/Park0WZ19} and SPADE~\cite{DBLP:conf/cvpr/Park0WZ19}) records consistently high IoU, accuracy, and FID scores. 
These scores indicate that the output images are aligned to the semantic masks relatively better but the image quality is lower, as we can see from the qualitative results.
The larger numbers of undetected faces (denoted as ``N/A'') also indicate low visual quality.
We confirmed that our pseudo labeling technique does not yield consistent improvements for the pixel-aligned approaches (indicated with ``*'').
In contrast, ours yields lower FID scores than the pixel-aligned approach and pSp~\cite{DBLP:journals/corr/abs-2008-00951} (even in the full-shot setting) consistently and is \chkB{overall} improved by increasing $N_l$ from 1 to 5.
Ours also outperforms pSp in the few-shot settings w.r.t. all the metrics except for N/A.}
The qualitative full-shot results are also included in the supplementary material.

\begin{figure}[t]
  \centering
  \includegraphics*[width=1.\linewidth, clip]{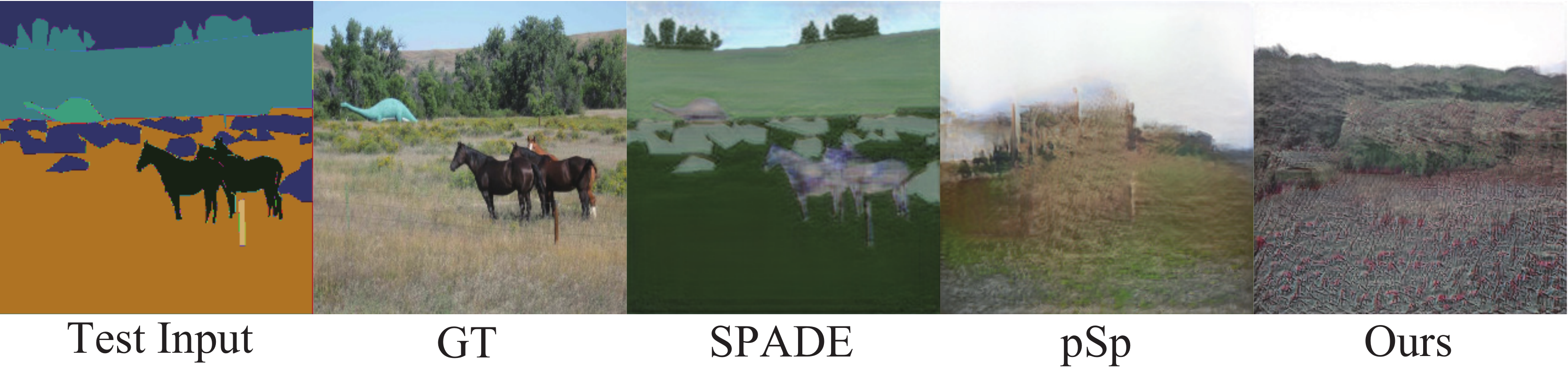}
   \caption{\chkE{Failure case with semantic classes that do not appear in few-shot training examples (``\textit{animal}'' in this case).}}

  \label{fig:limitation}
\end{figure}

\section{Discussion}
\chkA{Here we summarize the pros and cons of the related methods and ours. 
The pixel-aligned approach~\cite{DBLP:conf/cvpr/Park0WZ19} preserves spatial layouts specified by the semantic masks but fails to learn from our noisy pseudo labels due to the sensitivity to misaligned semantic masks.
Contrarily, \chkB{ours on top of pSp~\cite{DBLP:journals/corr/abs-2008-00951} is} tolerant of misalignment and thus \chkC{works} well with our pseudo labels. 
However, it is still challenging to reproduce detailed input layouts and to handle layouts that StyleGAN cannot generate.
A future direction is to overcome these limitations by, \chkC{e.g.,}} directly manipulating hidden units corresponding to semantics of input layouts~\cite{DBLP:conf/iclr/BauZSZTFT19}. 
\chkD{Another limitation is \chkE{that we cannot handle semantic classes unseen in the} few-shot examples. Figure~\ref{fig:limitation} shows \chkE{such an example} with \chkE{a} more challenging dataset, ADE20K~\cite{DBLP:conf/cvpr/ZhouZPFB017}.
Please refer to the supplementary material for more results.} 

\chkF{It is also worth mentioning that our method outperformed the full-shot version of pSp in FID.}
\chkG{This is presumably because our pseudo sampling could better explore StyleGAN's latent space defined by} \chkF{a large unlabeled dataset (e.g., 70K images in FFHQ and 48M images in LSUN church)} \chkG{than pSp, which} \chkF{uses limited labeled datasets} \chkG{for training the encoder.}

\section{Conclusion}

In this paper, we have proposed a simple \chkA{yet} effective method for few-shot semantic image synthesis \chkA{for the first time}. 
To compensate for the lack of pixel-wise annotation data, \chkA{we generate pseudo semantic masks via ($k$-)nearest-neighbor mapping between the feature vector of the pre-trained StyleGAN generator and each semantic class in the few-shot labeled data.}
\chkA{In each training iteration, we can generate a pseudo label from random noise to train an encoder~\cite{DBLP:journals/corr/abs-2008-00951} for controlling the pre-trained StyleGAN generator using a simple L2 loss.}
\chkA{The experiments with various datasets demonstrated that our method can synthesize higher-quality images with spatial control than competitive methods and works well even with sparse semantic masks such as scribbles and landmarks.}

{\small
\bibliographystyle{ieee_fullname}
\bibliography{egbib}
}

\newpage
\onecolumn
\appendix
\section{Implementation Details}
We implemented our method with PyTorch and ran our code on PCs equipped with GeForce GTX 1080 Ti. 
We used StyleGAN2~\cite{DBLP:conf/cvpr/KarrasLAHLA20} as a generator and pSp~\cite{DBLP:journals/corr/abs-2008-00951} as an encoder. 
We trained the encoder using the Ranger optimizer~\cite{DBLP:journals/corr/abs-2008-00951} with a learning rate of 0.0001. 
The batch size (i.e., the number of pseudo-labeled images per iteration) was set to 2.
We performed 100,000 iterations and took a day at most. 
\chkA{Regarding our multi-modal results, please refer to Section~\ref{sec:MultiModal} in this supplementary material.}

\section{Sparse Pseudo Labeling with Different Parameters}
\begin{figure}[h]
  \centering
  \includegraphics*[width=0.9\linewidth, clip]{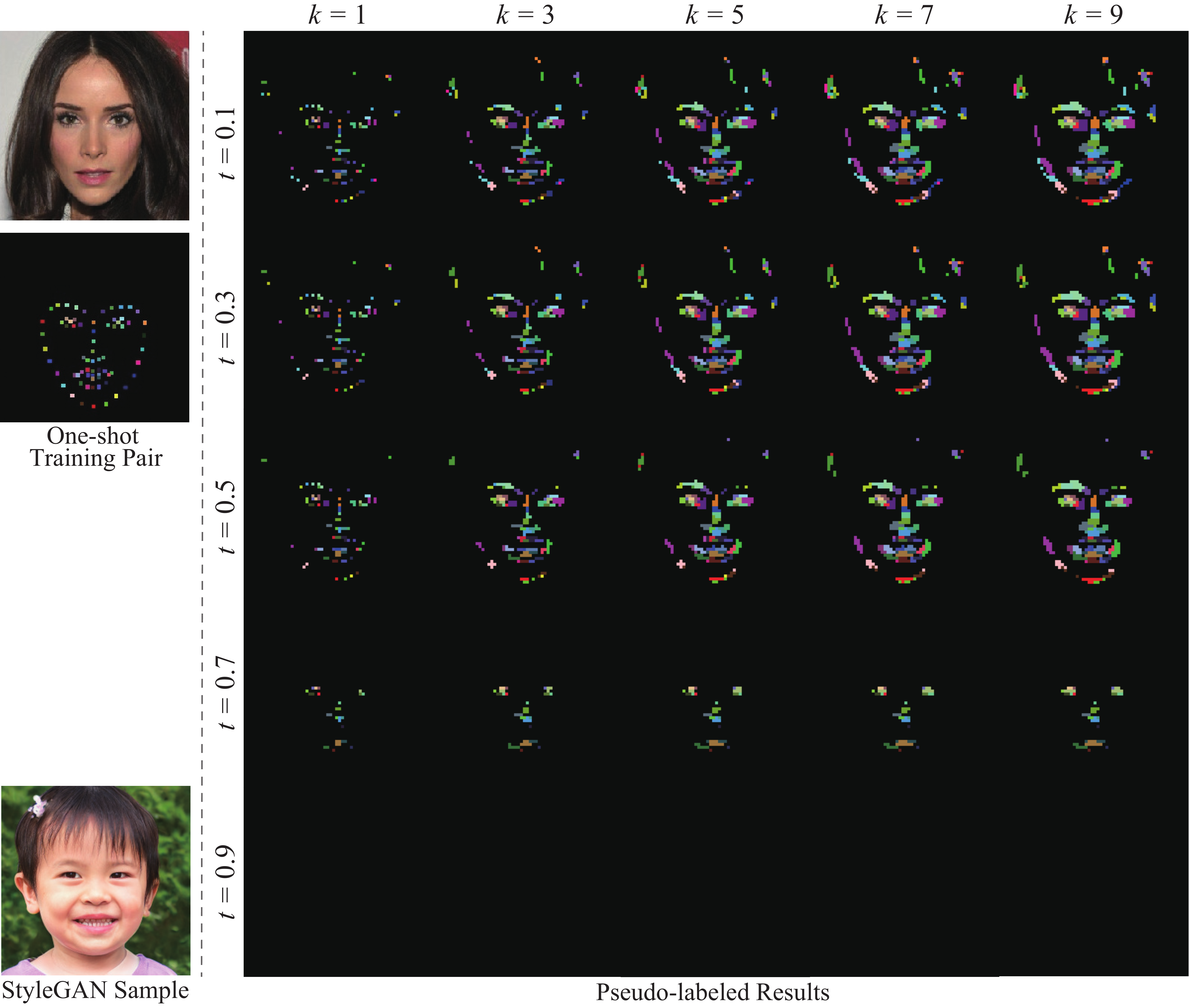}
  \caption{Sparsely pseudo-labeled results with different parameters $k$ and $l$. 
}
  \label{fig:kl}
\end{figure}

Figure~\ref{fig:kl} shows the sparsely pseudo-labeled results (right) for the StyleGAN sample (lower left) using different parameters $k$ and $t$ \chkA{with a one-shot training pair} (upper left).
\chkA{As explained in Subsubsection 3.3.2 in our paper, $k$ is used for top-$k$ matching between per-pixel feature vectors and representative vectors, whereas $t$ is a threshold of cosine similarity.}
\chkA{For all of our other results, we} set $k=3$ to reduce the number of misfetches of matched pixels and $t=0.5$ to reduce outliers. 

\newpage
\section{Images Reconstructed from Pseudo Semantic Masks During Training Procedure}
\begin{figure}[h]
  \centering
  \includegraphics*[width=0.98\linewidth, clip]{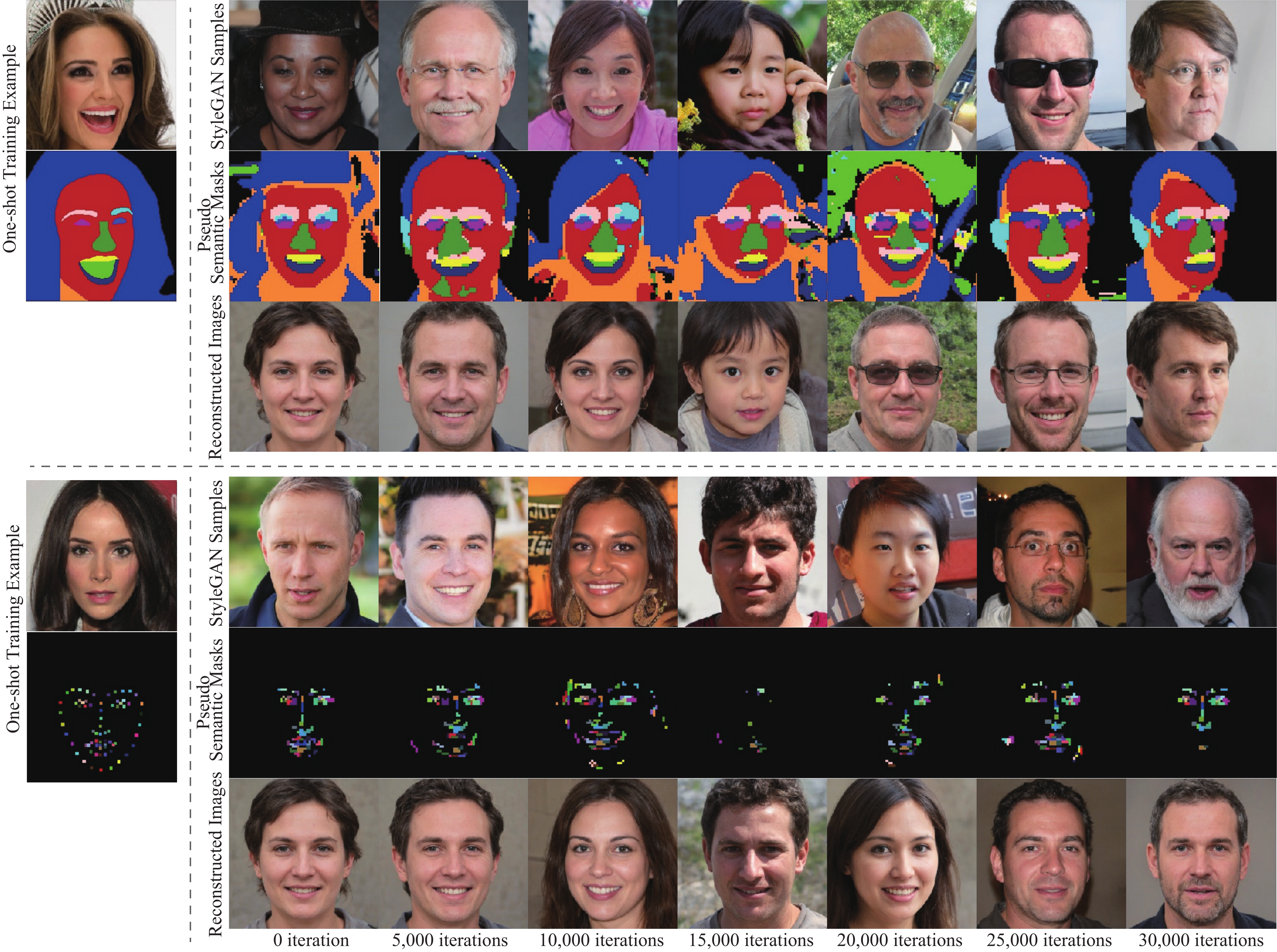}
  \caption{\chkA{Intermediate training outputs} with the StyleGAN pre-trained with the CelebA-HQ dataset. 
  \label{fig:t1}
}
\end{figure}

\begin{figure}[h]
  \centering
  \includegraphics*[width=0.98\linewidth, clip]{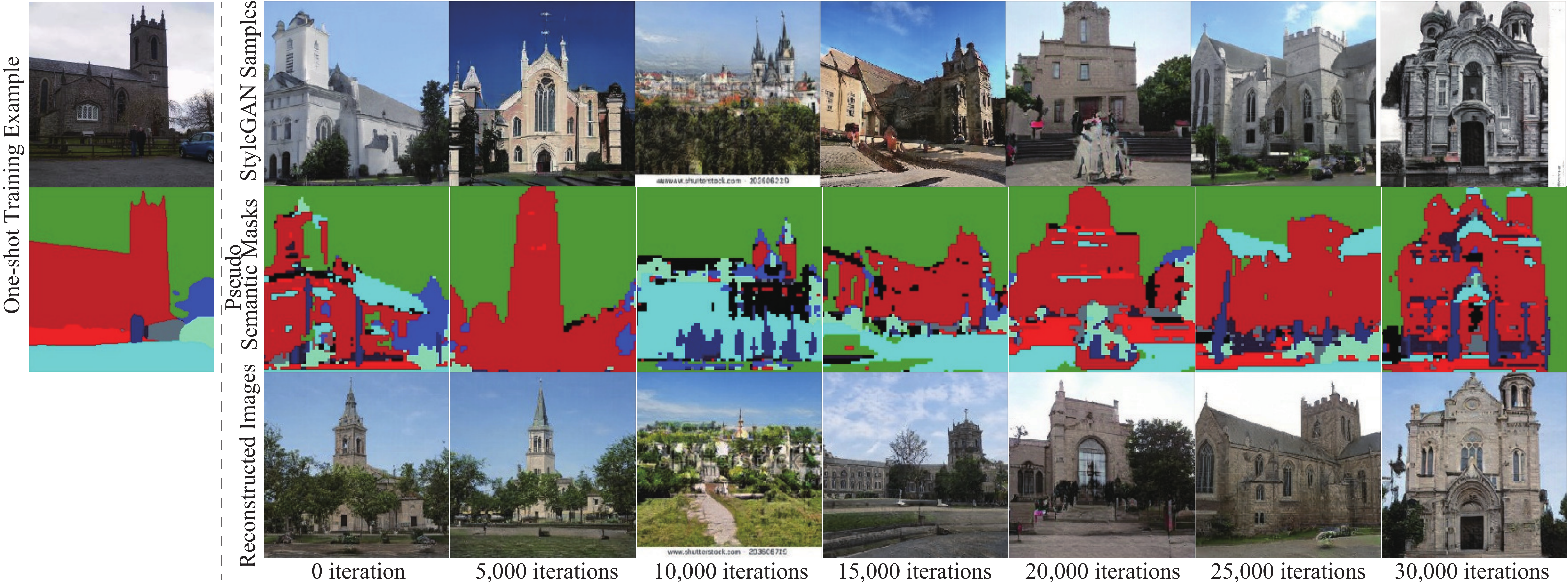}
  \caption{\chkA{Intermediate training outputs} with the StyleGAN pre-trained with the LSUN church dataset. 
   \label{fig:t2}
}
\end{figure}

\begin{figure}[h]
  \centering
  \includegraphics*[width=0.98\linewidth, clip]{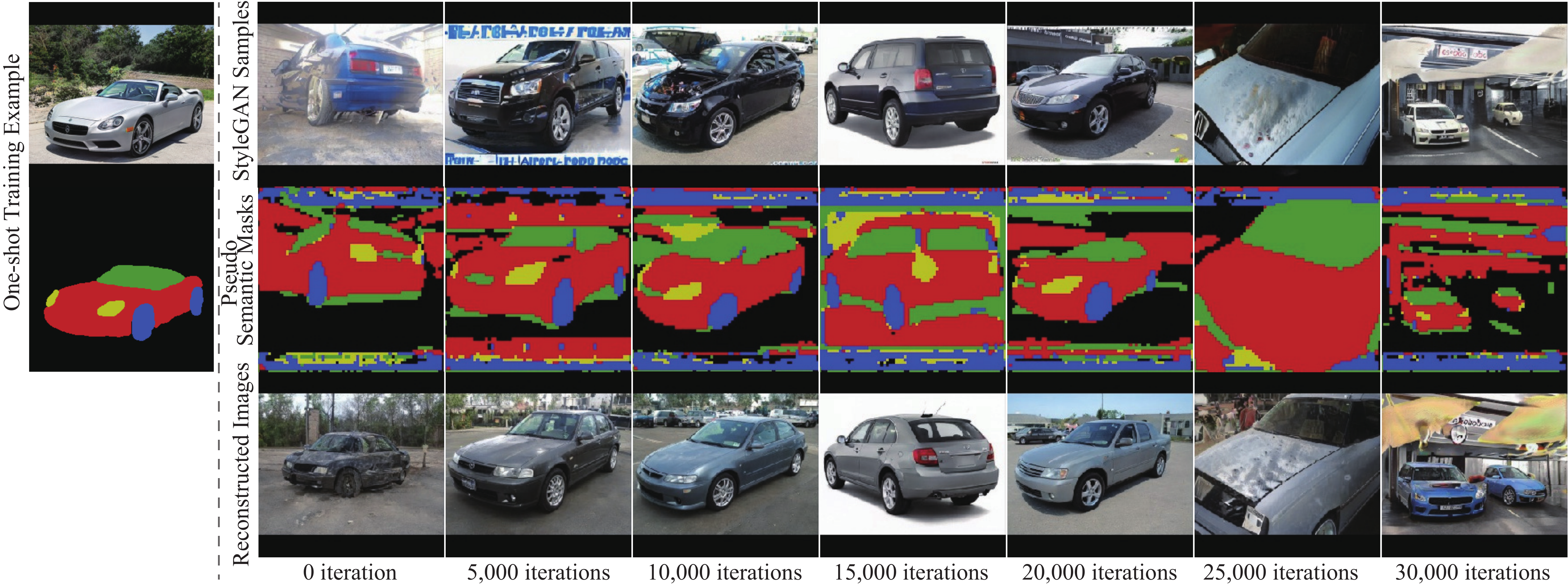}
  \caption{\chkA{Intermediate training outputs} with the StyleGAN pre-trained with the LSUN car dataset. 
   \label{fig:t3}
}
\end{figure}

\begin{figure}[h]
  \centering
  \includegraphics*[width=0.94\linewidth, clip]{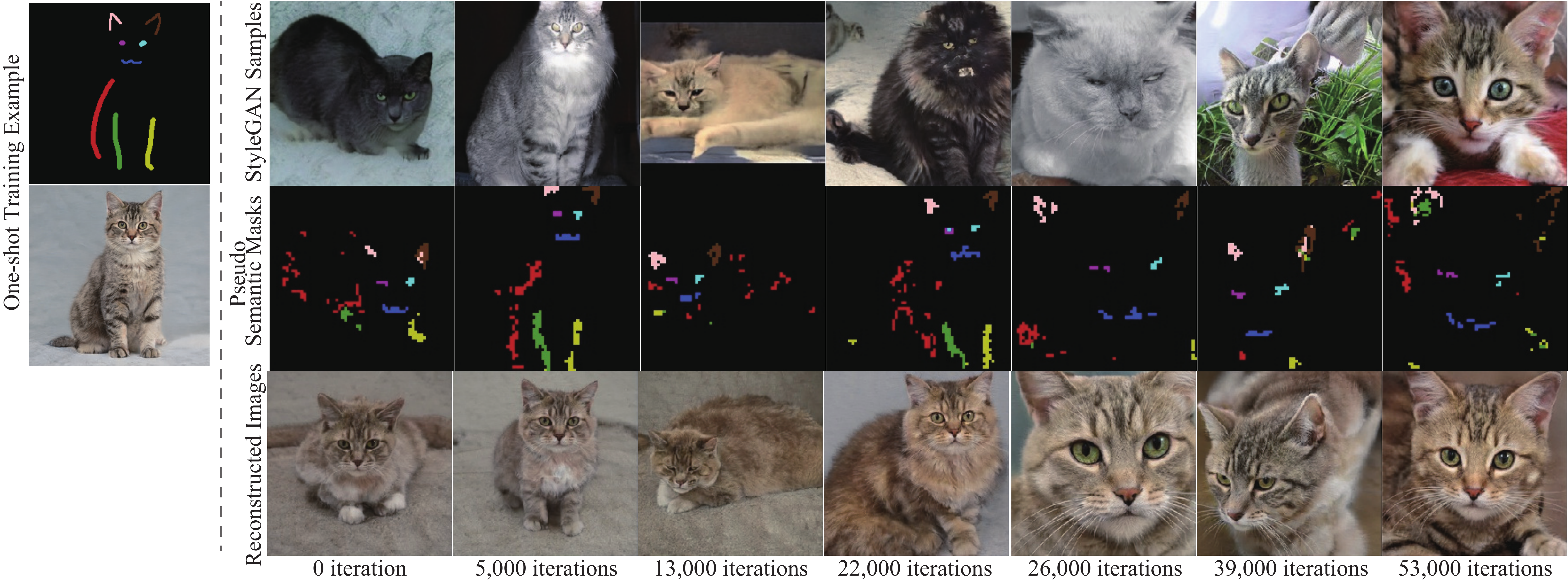}
  \caption{\chkA{Intermediate training outputs} with the StyleGAN pre-trained with the LSUN cat dataset. 
   \label{fig:t4}
}
\end{figure}

\begin{figure}[h]
  \centering
  \includegraphics*[width=0.94\linewidth, clip]{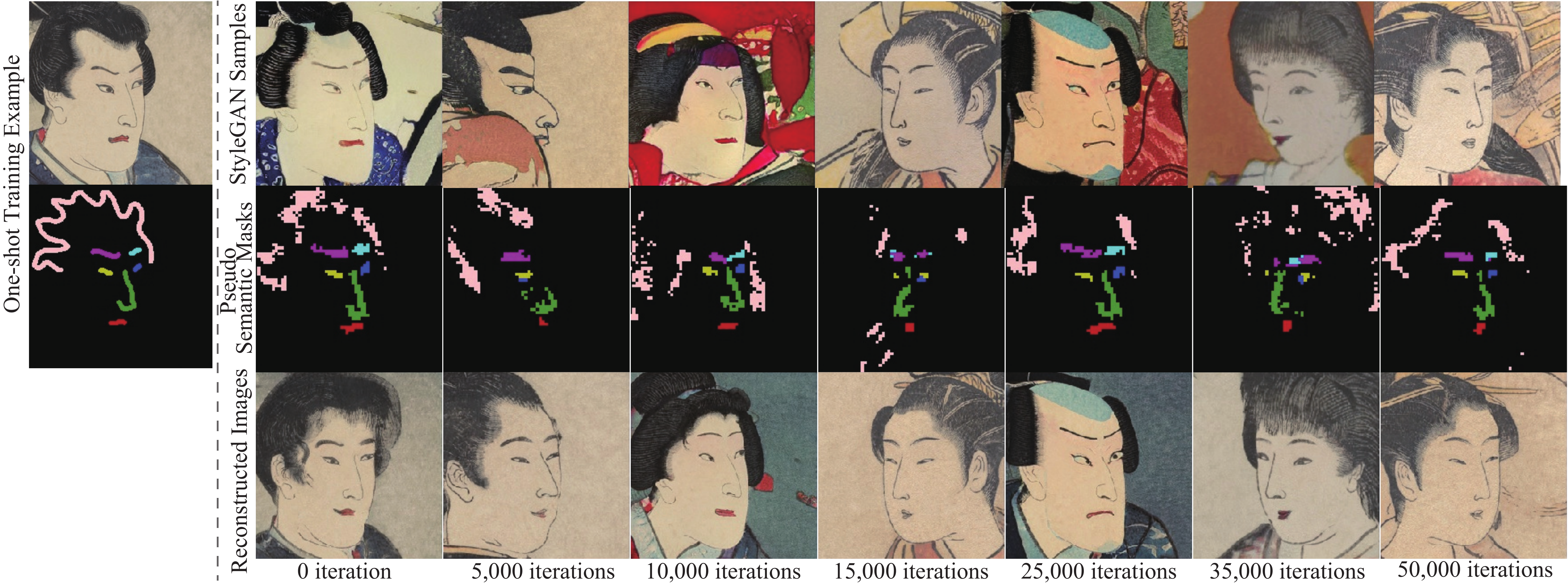}
  \caption{\chkA{Intermediate training outputs} with the StyleGAN pre-trained with the ukiyo-e dataset. 
   \label{fig:t5}
}
\end{figure}

\begin{figure}[h]
  \centering
  \includegraphics*[width=0.94\linewidth, clip]{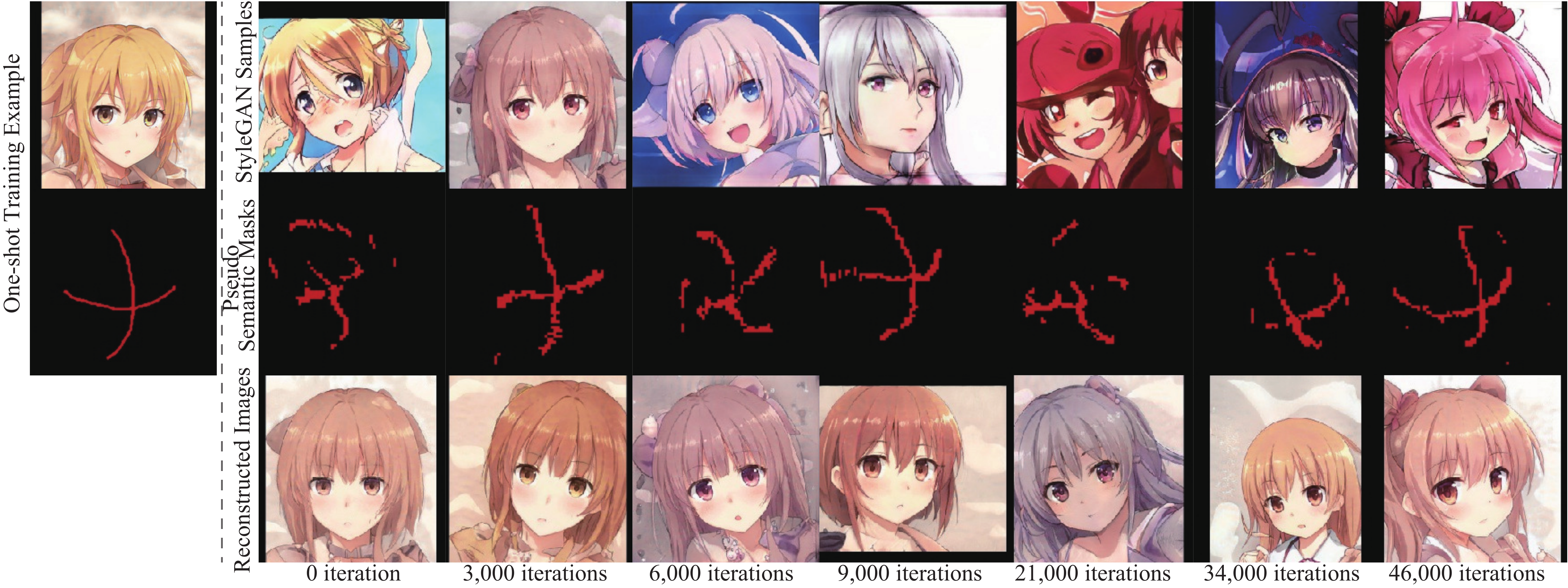}
  \caption{\chkA{Intermediate training outputs} with the StyleGAN pre-trained with the anime face dataset. 
   \label{fig:t6}
}
\end{figure}

Figures~\ref{fig:t1}, \ref{fig:t2}, \ref{fig:t3}, \ref{fig:t4}, \ref{fig:t5}, and \ref{fig:t6} show the \chkA{intermediate outputs in one-shot settings during training iterations, which is explained in Subsection 3.4 of our paper. 
For each set of results, we fed random noise vectors to the pre-trained StyleGAN generator to obtain synthetic images (top row) and feature vectors, from which we calculated pseudo semantic masks (middle row). We then used the pseudo masks to train the pSp encoder to generate latent codes for reconstructing images (bottom row).}
It can be seen that the layouts of the \chkA{bottom-row} images reconstructed from the \chkA{middle-row} pseudo semantic masks gradually become close to those of the \chkA{top-row} StyleGAN samples as the training iterations increase. 

\newpage
$ $
\newpage
\section{Multi-modal Results}
\label{sec:MultiModal}
\begin{figure}[h]
  \centering
  \includegraphics*[width=0.89\linewidth, clip]{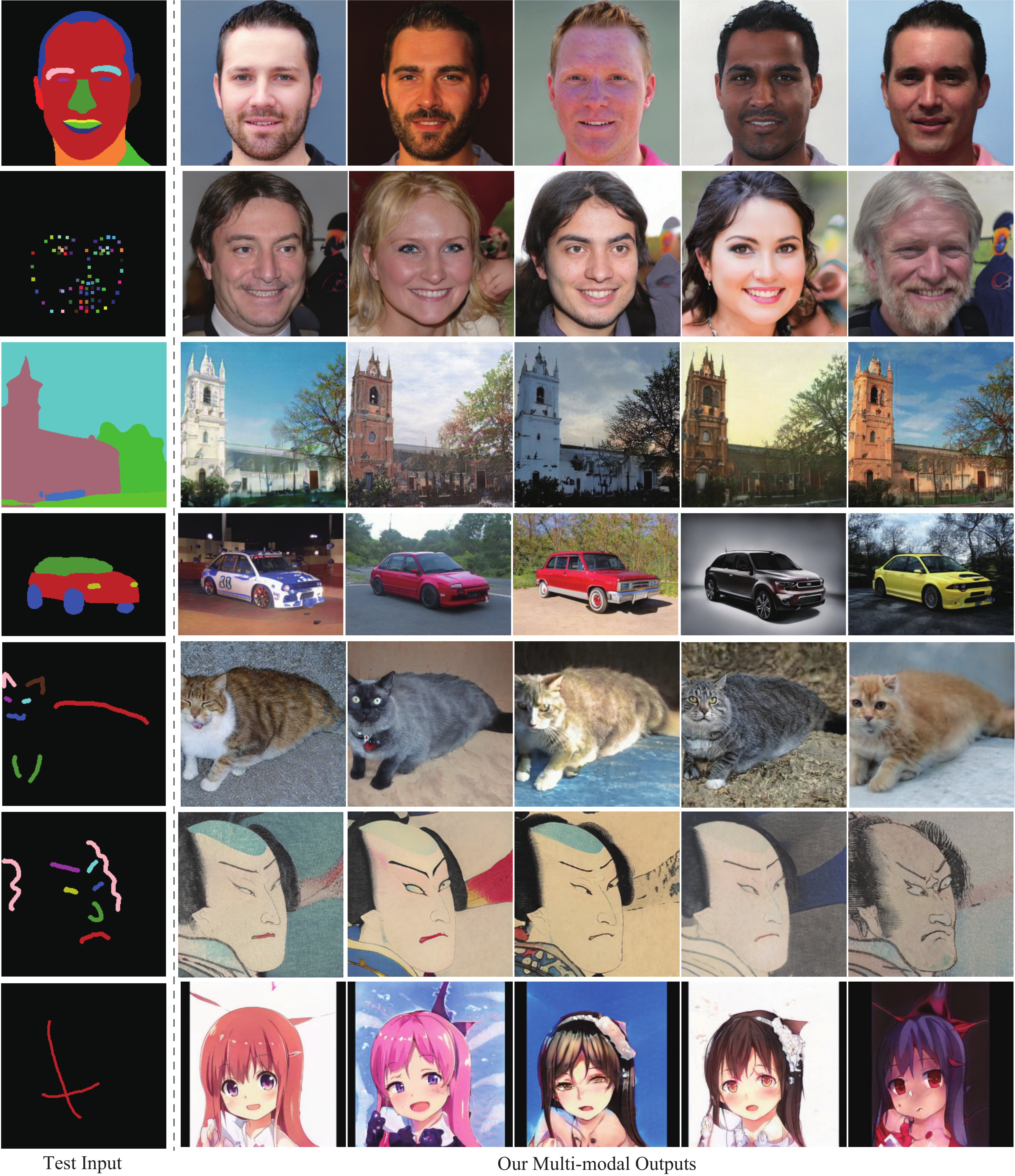}
  \caption{Multi-modal results of our method in few-shot settings.
}
\label{fig:multimodal}
\end{figure}

\chkA{Figure~\ref{fig:multimodal} demonstrates that} our method can generate multi-modal results.
To obtain multi-modal outputs in test time, we follow the same approach as pSp~\cite{DBLP:journals/corr/abs-2008-00951}; we fed latent codes encoded from an input layout to the first $l$ layers of the generator and random noise vectors to the other layers.
\chkA{While we used $l = 8$ for other results in our paper and this supplementary matrial, here we used different values of $l$ to create various outputs.
Specifically,} we set $l=$ 8, 5, 7, 5, 5, 5, and 5 from the top rows in Figure~\ref{fig:multimodal}.
\chkA{As explained in the pSp paper~\cite{DBLP:journals/corr/abs-2008-00951}, smaller $l$ affects coarser-scale styles whereas larger $l$ changes finer-scale ones.}

\newpage
\section{Additional Qualitative Results}
\begin{figure}[h]
  \centering
  \includegraphics*[width=0.96\linewidth, clip]{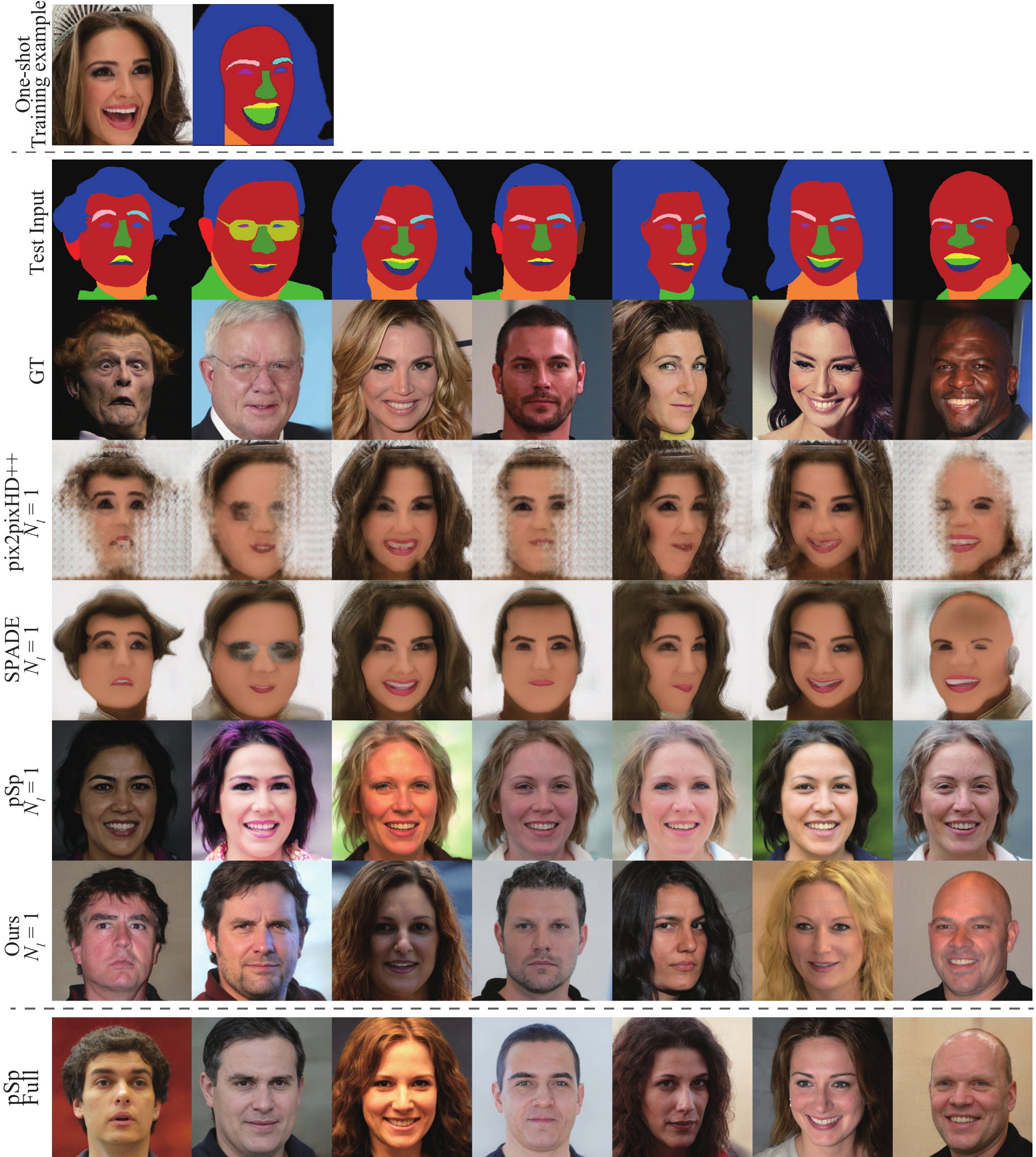}
  \caption{Additional visual comparison on the CelebAMask-HQ dataset. 
}
\label{fig:a1}
\end{figure}

\begin{figure}[h]
  \centering
  \includegraphics*[width=0.96\linewidth, clip]{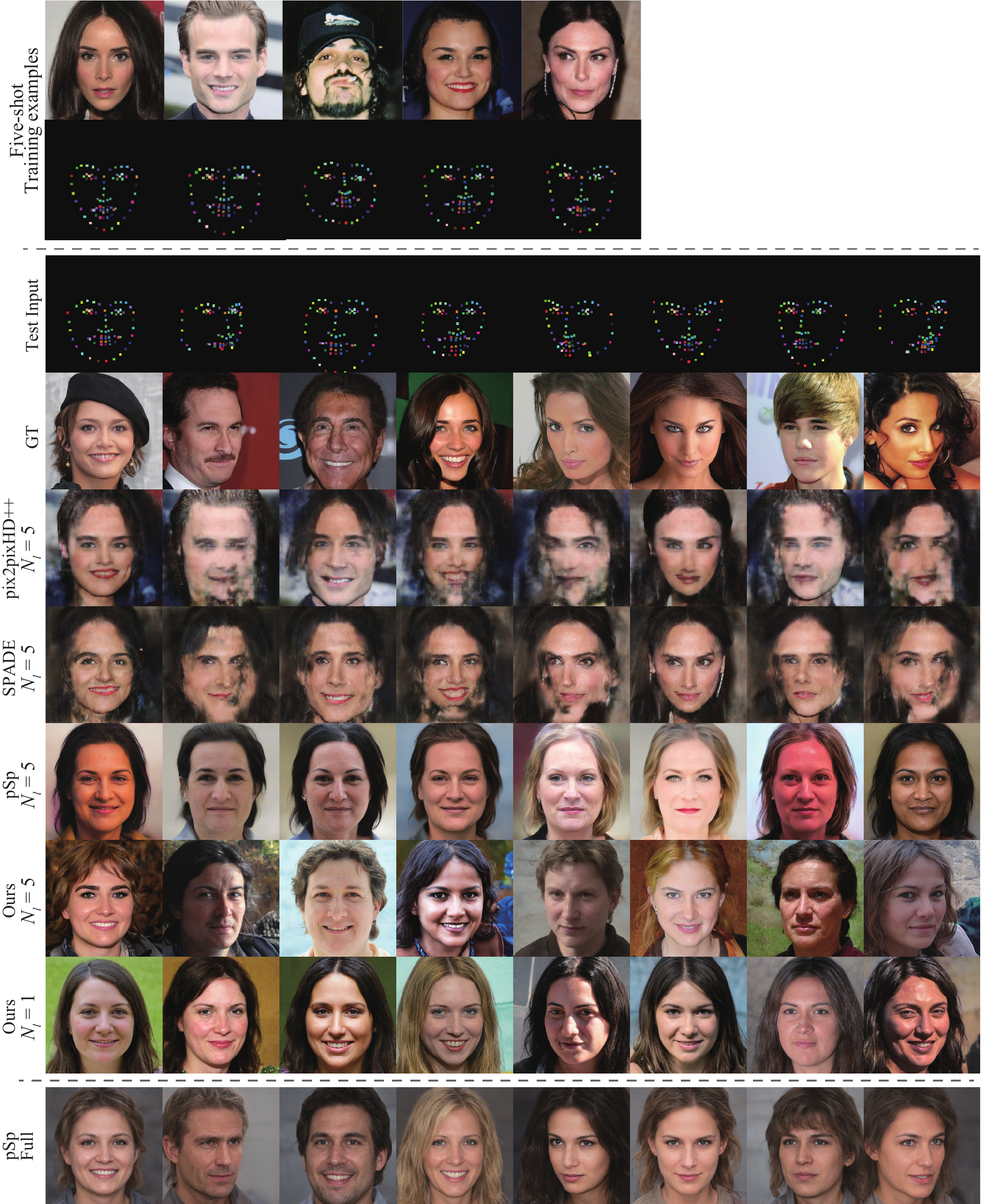}
  \caption{Additional visual comparison on the CelebALandmark-HQ dataset. 
}
\label{fig:a2}
\end{figure}

\begin{figure}[h]
  \centering
  \includegraphics*[width=0.96\linewidth, clip]{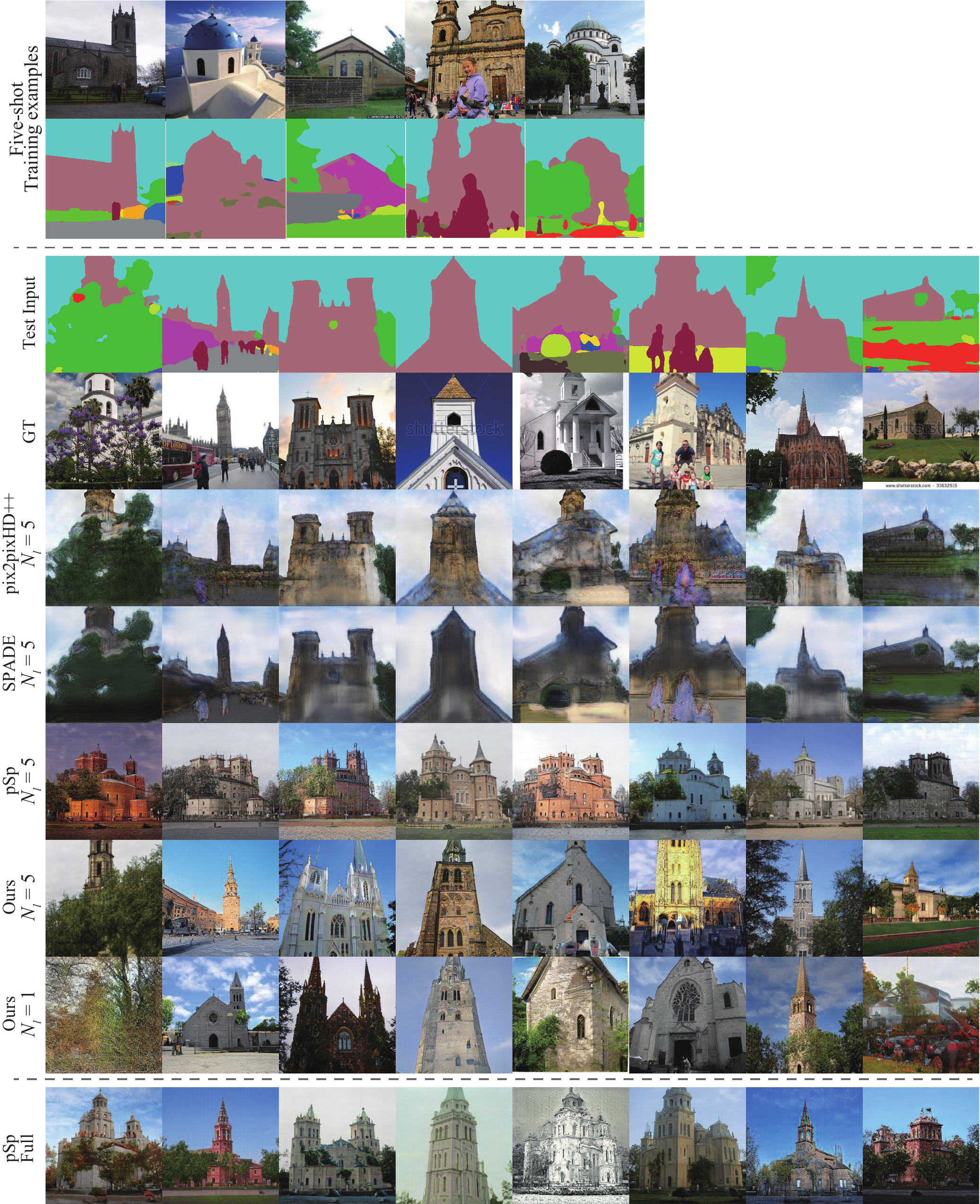}
  \caption{Additional visual comparison on the LSUN church dataset. 
}
\label{fig:a3}
\end{figure}

\begin{figure}[h]
  \centering
  \includegraphics*[width=0.96\linewidth, clip]{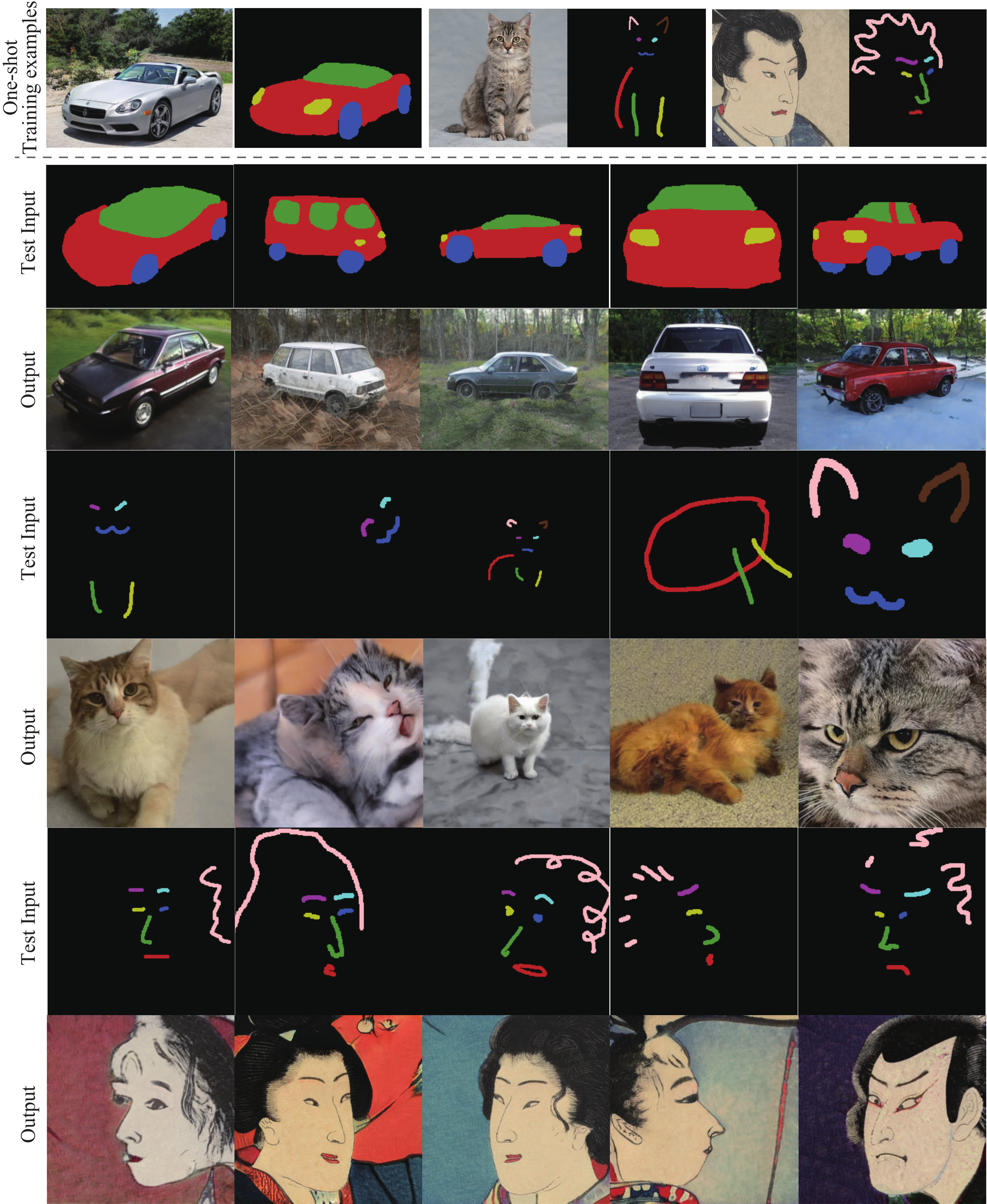}
  \caption{Our additional results obtained using various pre-trained StyleGANs in \chkA{one-shot settings}.}
\label{fig:a4}
\end{figure}

Figures~\ref{fig:a1}, \ref{fig:a2}, \ref{fig:a3}, and \ref{fig:a4} show the additional results. 
\chkA{The corresponding few}-shot training examples are the same as those shown in the paper. 
\newpage

\newpage

\newpage

\newpage
\section{Limitation}
\begin{figure}[h]
  \centering
  \includegraphics*[width=1.\linewidth, clip]{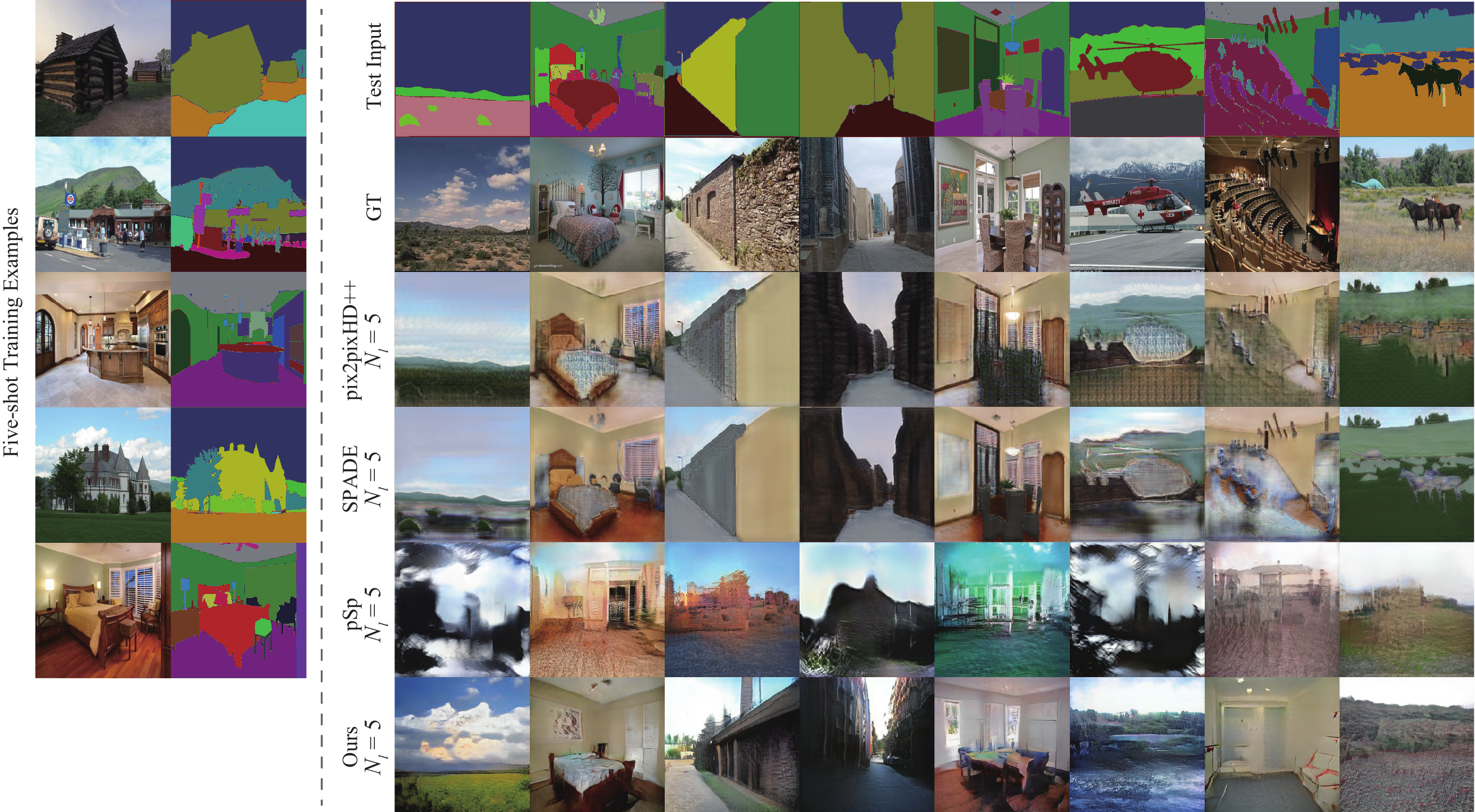}
  \caption{Qualitative comparison on the ADE20K dataset.
}
\label{fig:ade20k}
\end{figure}

Figure~\ref{fig:ade20k} and Table~\ref{tab:ade20k} show the results with a more challenging dataset, ADE20K~\cite{DBLP:conf/cvpr/ZhouZPFB017}, which consists of 20,210 training and 2,000 validation sets and contains indoor and outdoor scenes with 150 semantic classes. We used the training set without semantic masks to pre-train StyleGAN. Although our method can generate plausible images for some scenes, it struggles to handle complex scenes with diverse and unseen semantic classes. 

\begin{table}[h]
\centering
\caption{Quantitative comparison on the ADE20K dataset. }
\small
\begin{tabular}{r|c||c|c|c}
 \multicolumn{2}{c||}{} & \multicolumn{3}{c}{ADE20K}    \\ \cline{3-5} 
Method    & $N_l$       & fwIoU$\uparrow$          & accu$\uparrow$          & FID$\downarrow$    \\ \hline \hline
pix2pixHD++~\cite{DBLP:conf/cvpr/Park0WZ19} & $5$ &  39.2  & 56.0        & 110.0    \\
pix2pixHD++* & 5 & 18.7             &  31.5             & 142.8      \\
SPADE~\cite{DBLP:conf/cvpr/Park0WZ19} & 5 & 42.3  & 58.8         & 98.1       \\
SPADE* & 5 &  23.1             &  38.6             & 129.8         \\ \hline
pSp~\cite{DBLP:journals/corr/abs-2008-00951} & 1  & 6.3            & 17.9             & 187.5           \\
pSp~\cite{DBLP:journals/corr/abs-2008-00951} & 5    & 8.8             & 18.5             & 177.0     \\
Ours & 1  & 10.8            & 19.8            &155.4         \\
Ours & 5  & 15.8             & 28.3             & 95.1        \\ \hline
\end{tabular}
\label{tab:ade20k}
\end{table}

\end{document}